\documentclass[preprint,12pt]{elsarticle}
\usepackage[a4paper, margin=0.6in]{geometry}
\usepackage{amsmath,amsfonts}

\usepackage{algorithmic}
\usepackage{ragged2e}
\usepackage{textcomp}
\usepackage[OT1]{fontenc}
\usepackage{pifont}
\usepackage{multirow}
\usepackage{threeparttable}
\usepackage{caption}
\usepackage{amsmath}
\usepackage{booktabs}
\usepackage{comment}
\usepackage{dirtytalk}
\usepackage{float}
\usepackage{afterpage}
\usepackage[colorlinks=true]{hyperref}
\usepackage{tabularx}
\usepackage{makecell}
\usepackage{adjustbox}
\usepackage{enumitem}
\usepackage{url}
\usepackage{booktabs}
\usepackage{multirow}
\usepackage{siunitx}
\usepackage{lipsum}
\usepackage{mathtools}

\usepackage{textcomp}
\usepackage{microtype}
\usepackage{booktabs,makecell,tabularx}
\usepackage{subfigure}
\usepackage{tabularray}
\usepackage{microtype}

\newcolumntype{d}{>{\hsize=0.15\hsize}X}
\newcolumntype{D}{>{\hsize=0.2\hsize}X}
\newcolumntype{h}{>{\hsize=0.35\hsize}X}
\newcolumntype{a}{>{\hsize=0.75\hsize}X}
\usepackage{xcolor}
\usepackage{graphicx}
\usepackage{pdflscape}
\usepackage{soul}
\setul{0.1ex}{0.3ex}

\newcommand{\mul}[2]{{\setulcolor{#1}\ul{\texttt{#2}}\hspace{-0.001em}}}

\usepackage[most]{tcolorbox}

\usepackage{tikz}
\newcommand{\cvtag}[2]{
  \tikz[baseline]\node[anchor=base,fill=#1,rounded corners,inner xsep=1ex,inner ysep =0.5ex,text height=1ex,text depth=.15ex]{{\scriptsize \texttt{#2}}};
}

\definecolor{ifc}{HTML}{A7CBFF}
\definecolor{sar}{HTML}{B6DC97}
\definecolor{icb}{HTML}{8AD5D4}
\definecolor{s4b}{HTML}{E7F3DC}
\definecolor{s4e}{HTML}{93BA70}
\definecolor{s4s}{HTML}{CEE8BA}
\definecolor{bot}{HTML}{8380A5}
\definecolor{sos}{HTML}{FCE0DC}
\definecolor{kpi}{HTML}{C9C9C9}
\definecolor{bri}{HTML}{F5A197}
\definecolor{oth}{HTML}{EDEDED}
\definecolor{sbm}{HTML}{D8F1F1}
\definecolor{ssn}{HTML}{DAA283}
\definecolor{phy}{HTML}{ACAAC3}
\definecolor{oeccomplete}{HTML}{D4AF37}
\definecolor{oecpartial}{HTML}{C0C0C0}

\definecolor{x}{HTML}{FEFEFE}
\definecolor{bd}{HTML}{318886}
\definecolor{bl}{HTML}{D8F1F1}
\definecolor{lks}{HTML}{7C7C7C}

\newcommand{\ftag}[1]{
 \tikz[baseline]\node[anchor=base,fill=bd, rounded corners,inner xsep=0.3em,inner ysep =0.4em,text height=0.6em]{{\footnotesize{\color{bl}\textbf{\textsf{#1}}}}};
}

\newcommand{\changeurlcolor}[1]{\hypersetup{linkcolor=#1,urlcolor=#1}}

\definecolor{rd}{HTML}{C42F1A}
\definecolor{rl}{HTML}{F8D2CC}
\newcommand{\rtag}[1]{
 \tikz[baseline]\node[anchor=base,fill=rd, rounded corners,inner xsep=0.3em,inner ysep =0.4em,text height=0.6em]{{\footnotesize{\color{rl}\textbf{\textsf{#1}}}}};
}

\usepackage{tabularx,ragged2e,siunitx}
\newcolumntype{Y}[1]{>{\Centering\hspace{0pt}\hsize=#1\hsize}X}

\usepackage{amssymb}

\journal{}

\begin{document}

\begin{frontmatter}

\def\tsc#1{\csdef{#1}{\textsc{\lowercase{#1}}\xspace}}
\tsc{WGM}
\tsc{QE}
\pagenumbering{arabic}
\title{A Survey on Semantic Modelling for Building Energy Management}

\ead{\{maniakor, vvcogo, pmf\}@ciencias.ulisboa.pt}
\author{Miracle E. Aniakor\textsuperscript{\hyperlink{corrauth}{*}}, Vinicius V. Cogo, Pedro M. Ferreira}
\address{LASIGE, DI, Faculdade de Ciências, Universidade de Lisboa, Portugal}

\cortext[1]{Corresponding author at LASIGE, DI, Faculdade de Ciências, Universidade de Lisboa, Portugal.}

\begin{abstract}
Building Energy Management (BEM) is central to reducing energy use and CO$_2$ emissions in the building sector. Although Internet of Things (IoT) technologies now provide extensive operational data, heterogeneous data models, device descriptions, and contextual representations continue to limit semantic interoperability. These limitations hinder the development of generalisable, autonomous, and context-aware BEM applications. Ontologies address this challenge by providing structured, machine-interpretable representations of building data, systems, and operational context.
This survey examines semantic modelling for BEM during the building operational phase. It reviews sixty-one semantic models and analyses more than twenty ontology-based BEM use cases reported in individual studies, distinguishing single-ontology, two-ontology, and multiple-ontology modelling strategies. It further quantifies Ontology Instantiation Rates (OIR) and necessity to extend (NTE) across the reviewed use cases. To support evidence-based assessment of ontology use, the survey introduces the notion of Ontology Evidence Completeness (OEC), a measure of whether studies explicitly map operational concepts to the ontology classes used to represent them.
The findings show that current semantic models more consistently represent physical building structure, technical systems, sensing devices, and observable operational data than abstract and dynamic operational concepts, although these physical concepts may still need to be adapted or combined with other ontologies in specific application contexts. Concepts such as key performance indicators, assessments, services, control logic, optimisation tasks, and computational workflows remain less consistently covered. Applied BEM studies therefore frequently depend on ontology reuse, integration, specialisation, external inheritance, or application-specific extension to address coverage and interoperability gaps across both physical and operational aspects of BEM. By synthesising these patterns, this survey clarifies the capabilities and limitations of existing semantic models and identifies directions for more interoperable, generalisable, and context-aware BEM systems.
\end{abstract}


\end{frontmatter}

\section{Introduction}
\label{sec:Introduction}

The building sector remains among the most energy-intensive domains in the European Union (EU), accounting for approximately $40\%$ of total energy use and $36\%$ of CO$_2$ emissions~\cite{eucommission}. Improving building energy performance is therefore central to reducing resource consumption and carbon emissions. Building Energy Management (BEM) addresses this objective through strategies and systems that monitor, analyse, and control energy use while also supporting occupant comfort, energy flexibility, predictive analysis, maintenance, and fault detection.

\begin{figure}[!t]
\centering
\begin{minipage}{0.40\textwidth}
\subfigure[Relationship among the buildings' lifecycle phases and examples of the respective (static or dynamic) data they manage.]{
        \includegraphics[width=1\columnwidth]{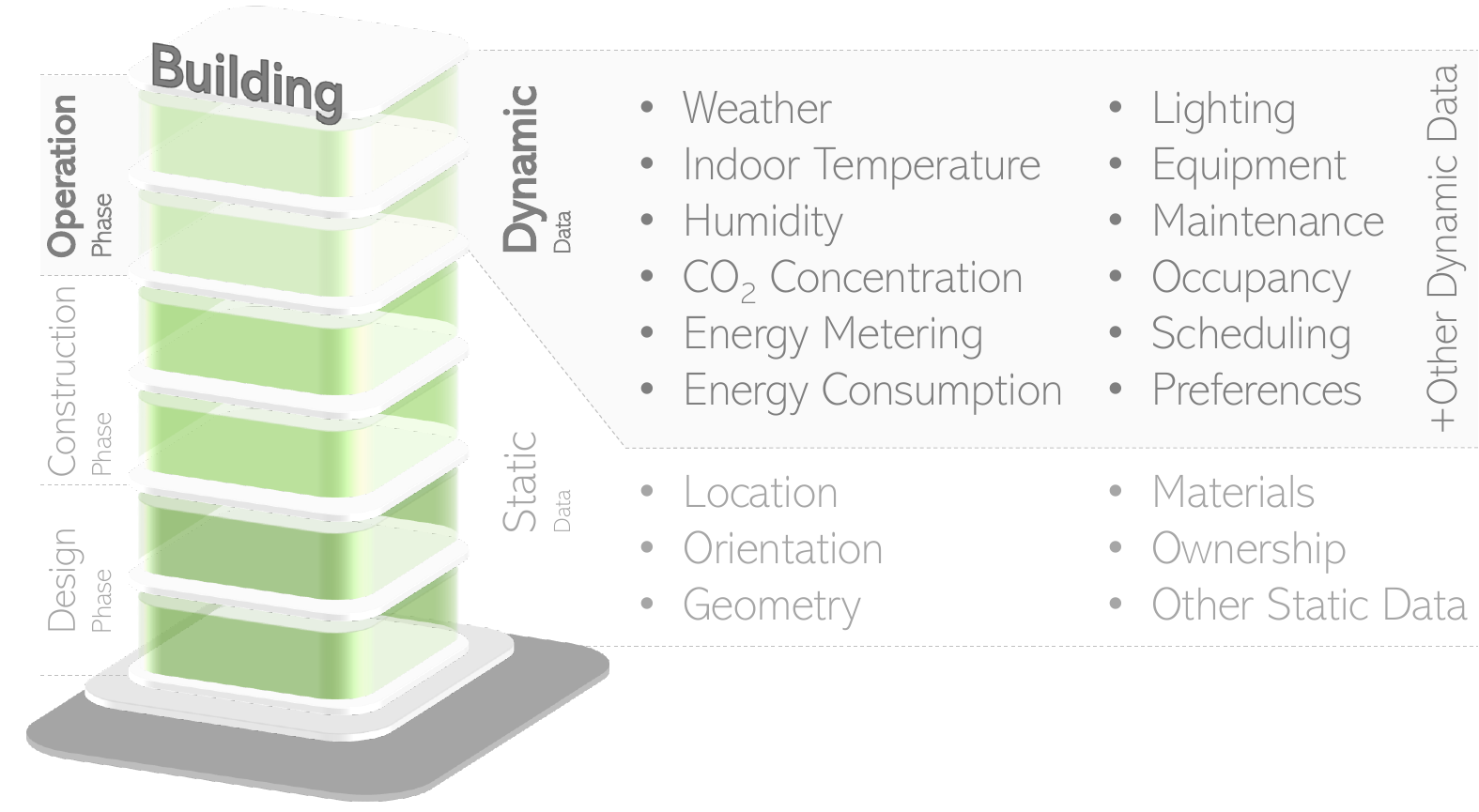}
        \label{fig:Static data}
}
\end{minipage}
\hspace{0.03\textwidth}
\begin{minipage}{0.40\textwidth}
\centering
\subfigure[Steps for intelligent context-aware decision support systems.]{
	\includegraphics[width=1\columnwidth]{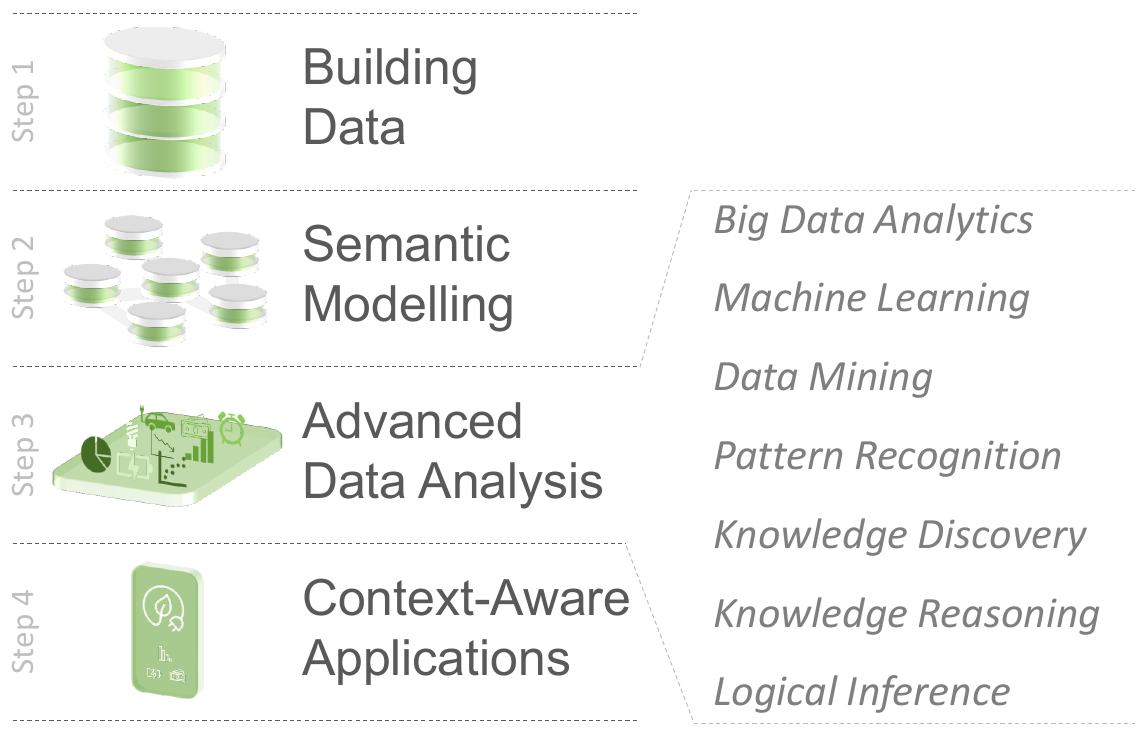}
	\label{fig:SDSS}
}
\end{minipage}

\caption{BEM Applications Framework}
\label{fig:xxx1}
\end{figure}

The operational phase of a building is especially important for BEM because it deals with continuously changing conditions and data streams from building systems, sensors, actuators, occupants, weather services, and energy resources.
Unlike the design and construction phases, which mainly provide static building information, operation requires dynamic data to infer building state and support timely decisions~\cite{ramesh2010life}. Figure~\ref{fig:Static data} illustrates this distinction, while Figure~\ref{fig:SDSS} summarises how contextual information can support intelligent decision-making in BEM. 
Figure~\ref{fig:SDSS} delineates this process as a sequence of steps from building data and enriched semantic modelling, through advanced data analytics that provide actionable operational knowledge, to context-aware applications.

The increasing adoption of Internet of Things (IoT) technologies has expanded the range of devices and information sources available to BEM systems. However, transferring applications across buildings remains difficult because data are represented differently by devices, vendors, protocols, software tools, and building management systems. This heterogeneity creates semantic interoperability challenges and limits the development of BEM applications that are generalisable, autonomous, and context-aware~\cite{hong2009context}.

Semantic modelling addresses this problem by providing machine-interpretable descriptions of building entities and their relationships.
In building operations, semantic models can represent entities such as spaces, zones, equipment, sensors, actuators, measurements, control points, and contextual data in a shareable form.
This semantic layer supports reusable queries, reasoning, data integration, and the development of more portable BEM applications.
When combined with building cyber-physical systems and Artificial Intelligence (AI), semantic models can help context-aware applications interpret current building conditions and support assessment, control, optimisation, and other operational decision-making tasks~\cite{xiao2014data,lazarova2016fault}.

Within this semantic modelling landscape, many ontological efforts have emerged to support interoperability in buildings and related IoT systems, from which the following have achieved more prominent utilisation.
The Building Topology Ontology~(BOT)~\cite{botOntology} supports building topology and spatial structure, the Semantic Sensor Network and Sensor, Observation, Sample, and Actuator ontologies (SSN/SOSA)~\cite{compton2012ssn,janowicz2019sosa} represent sensors, observations, actuators, and procedures, the Smart Applications REFerence ontology (SAREF) and its extensions~\cite{daniele2015created,sarefcore} support interoperability across IoT, device, building, and energy-domain, and Brick~\cite{balaji2016brick} provides a detailed schema for building equipment, points, locations, and operational metadata. While these models demonstrate substantial progress, their differing modelling scopes and varying levels of representational depth help explain why semantic coverage in BEM use cases remains uneven across physical building entities, sensing infrastructure, computational processes, and decision-support concepts.

Existing semantic models do not provide complete coverage of the concepts required for intelligent context-aware BEM. They generally provide more mature coverage of physical building entities, including topology~\cite{rasmussen2021bot}, devices~\cite{sarefcore}, equipment~\cite{balaji2018brick}, and sensors and measurements~\cite{haller2019modular,pritoni2021metadata}, than of abstract operational concepts such as key performance indicators (KPIs)~\cite{li2019enhancing,zhang2021linking}, assessments~\cite{corry2015performance}, services, control logic~\cite{schneider2017ontology}, optimisation objectives~\cite{thomsen2025ontology}, schedules~\cite{saref4ener}, and the associated computational workflows~\cite{yoon2025ontology}.
Such concepts are essential because intelligent BEM depends not only on describing building assets, but also on representing how operational data are evaluated and how decisions are triggered.

This limitation is particularly visible in the functional architecture of BEM systems, which can be understood through three interconnected layers: \emph{KPIs}, \emph{assessments}, and \emph{services}~\cite{lork2019ontology}. KPIs provide measurable indicators of building performance, assessments interpret those indicators against operational objectives, and services trigger actions intended to improve performance. Existing work has modelled some of these concepts independently, for example through KPI ontologies~\cite{kpiOntology}. However, modelling them in isolation limits their usefulness for automated decision support because the relationships between performance metrics, evaluations, and responsive actions are often not specified clearly enough.

Several surveys have examined semantic models in the building domain. Qiang \emph{et al.}~\cite{qiang2023systematic} examined four prominent building ontologies to support semantic interoperability and reusability in one real-world application, while Esnaola \emph{et al.}~\cite{esnaola2020ontologies} focused specifically on sensors and actuators.
Other reviews~\cite{fierro2022survey, lygerakis2022knowledge} discuss knowledge graphs and metadata schemas for smart buildings, but provide limited analysis of semantic coverage across diverse operational BEM applications and use cases.
This work goes beyond these studies by surveying a broader range of available semantic models, ontologies, and use cases, analysing how semantic models are applied in building operations, and tracing the extent to which task-relevant BEM concepts are explicitly represented in reviewed use cases.
Rather than focusing only on the availability or general scope of semantic models, the work examines which operational concepts are documented at the class level, which concepts remain unsupported or insufficiently specified, and which modelling strategies are used to address these gaps. In doing so, the survey provides a broader assessment of semantic coverage, ontology reuse, integration, extension practices, and evidence traceability in operational BEM applications.
This approach is important because mentioning an ontology in a use case does not necessarily show that the ontology explicitly covers the concepts required by the task.
A use case may rely on implicit assumptions, undocumented mappings, application-specific extensions, or combinations of several ontologies. Therefore, evaluating semantic models for BEM requires attention to use-case-level evidence: which concepts are required, which ontology classes are used, which concepts are missing, and which modelling strategies were adopted to address missing concepts.

To address the shortcomings of previous surveys, this work focuses on semantic modelling for BEM during the building operational phase. It does not propose a new ontology or implement a context-aware BEM system. Instead, it establishes where existing semantic models succeed, where they remain limited, and how these limitations are handled in applied BEM studies. The survey makes the following contributions:

\begin{enumerate}
    \item A review of sixty-one semantic models relevant to building operations and a characterisation of their scope, main concepts, and limitations.
    \item An analysis of concrete BEM use cases according to single-ontology, two-ontology, and multiple-ontology modelling strategies.
    \item Proposing the Ontology Evidence Completeness (OEC) notion, an evidence-oriented assessment concept introduced to evaluate whether ontology-based BEM studies provide traceable concept-to-class mappings for task-relevant operational concepts.
    \item A quantitative synthesis of ontology instantiation rate (OIR), necessity to extend (NTE), and the modelling strategies used to address semantic gaps, including class specialisation, external inheritance, reuse, integration, and application-specific extension.
    \item The identification of recurring research gaps in semantic modelling for BEM, particularly the representation of abstract computational processes, ontology extension practices, and energy flexibility resources.
\end{enumerate}

Guided by these contributions, the survey addresses the following research questions:

\begin{enumerate}
    \item[] \emph{RQ1}. Why are ontologies needed in building operation applications?
    \item[] \emph{RQ2}. Which ontologies and semantic models have been used in the building operation phase?
    \item[] \emph{RQ3}. Which building operational use cases have applied these ontologies?
    \item[] \emph{RQ4}. What limitations arise when applying existing ontologies to building operations and BEM applications?
\end{enumerate}

The remainder of the paper is organised as follows.
Section~\ref{sec:research methodology} presents the review methodology employed in this survey.
Section~\ref{sec:context_bkg} introduces the technical background and core concepts for operational BEM ontologies and explains the role of semantic modelling in building operations, answering~\emph{RQ1}.
Section~\ref{sec:semantic models} reviews the semantic models used in the domain, answering \emph{RQ2}.
Section~\ref{sec:usecases} analyses concrete ontology use cases and quantitative coverage trends, answering \emph{RQ3}.
Section~\ref{sec:Discussion} discusses the limitations, trends, and future directions identified from the surveyed studies, addressing~\emph{RQ4}.
Finally, Section~\ref{sec:conclusion-future} concludes the survey.

\section{Research methodology}
\label{sec:research methodology}

\begin{figure*}[!t]
    \centering
    \includegraphics[width=1.00\textwidth]{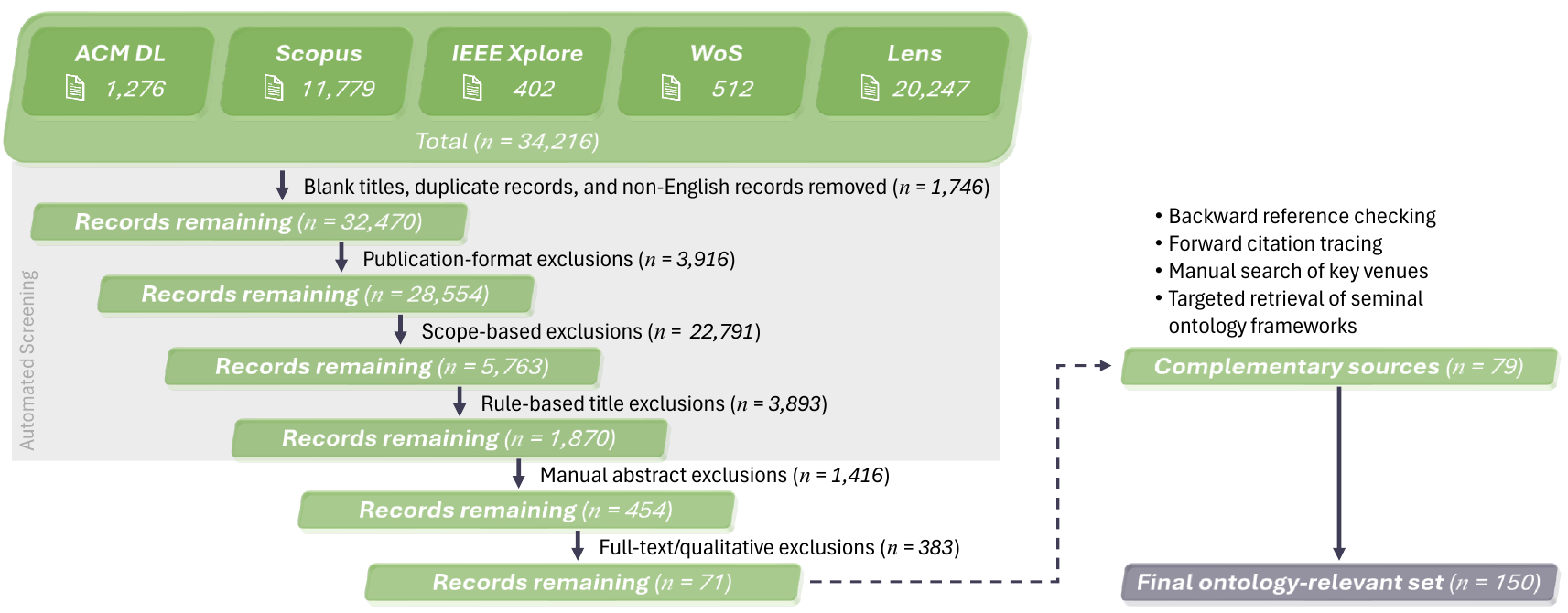}
    \caption{PRISMA-oriented study selection and complementary source-identification workflow.}
    \label{fig:study-selection}
\end{figure*}

This section describes the selection procedure used to construct the corpus analysed in relation to the research questions introduced in Section~\ref{sec:Introduction}. The review adopts a hybrid strategy that combines a PRISMA-oriented database-search workflow~\cite{page2021prisma} with complementary source-identification procedures, as shown in Figure~\ref{fig:study-selection}.
This strategy was necessary because ontology-centred surveys rely on two types of evidence: peer-reviewed studies that report ontology use in building operational applications, and technical sources that document the ontologies, standards, serialisations, and specifications needed to interpret those studies.
A database-only workflow would provide a reproducible peer-reviewed corpus but could risk excluding ontology documentation, standards materials, and project specifications that are essential for assessing semantic scope and technical implementation. Conversely, relying only on complementary source identification would reduce reproducibility.
The hybrid design, therefore, separates the peer-reviewed analytical corpus from support technical sources while making both identification routes explicit.

The bibliography includes both the ontology-relevant review corpus and supporting references used for background, methodology, and contextual discussion. A total of 150 references were classified as ontology-relevant to the scope of the review.
Within this ontology-relevant subset, 109 journal, conference, and workshop papers formed the peer-reviewed analytical corpus.
The remaining 41 are technical and contextual support sources, including technical reports, online ontology specifications, books, and thesis material, used to provide standards context, ontology documentation, and domain-specific technical grounding.

\subsection{Review design and search strategy}
\label{sec:review_design}

The database-search workflow used a structured keyword search to retrieve sources related to semantic modelling, ontologies, linked data, knowledge graphs, metadata models, and building energy management. The refined search string was:

{\footnotesize
\begin{center}
``\texttt{("Semantic modelling" OR "Semantic modeling" OR "Ontology" OR "Ontological Framework" OR "Data Model" OR "Knowledge Graph" OR "Linked Data" OR "Semantic") \\
\textbf{AND} (("Building" AND "Energy Management") OR ("Building" AND "Energy Management System") OR "BEM" OR ("KPI" AND "Building") OR ("Key Performance Indicator" AND "Building"))}''
\end{center}}

The search was queried in the five databases presented in Figure~\ref{fig:study-selection}: ACM Digital Library (DL, \url{https://dl.acm.org}), IEEE Xplore (\url{https://ieeexplore.ieee.org}), Lens (\url{https://lens.org/}), Scopus (\url{https://scopus.com}), and Web of Science (WoS, \url{https://webofscience.com}).
In total, 34,216 records were retrieved, comprising 1,276 records from ACM DL, 402 from IEEE Xplore, 20,247 from Lens, 11,779 from Scopus, and $512$ from WoS.
Duplicate records were expected in this initial merged dataset since the same publication may be indexed in multiple databases.

Venue-level bibliometric indicators, such as journal quartile, impact factor, or h-index, were not used as inclusion criteria because such metadata were not uniformly available across the corpus and may not reflect the relevance of ontology specifications or standards-related contributions.
Selection was therefore based on scope alignment, conceptual relevance, publication type, full-text availability, and the presence of an explicit ontology-based semantic contribution. This was important because influential ontology resources may appear outside high-impact journal venues.
For example, the SAREF ontology~\cite{daniele2015created} has made practical contributions to widely adopted technologies and has become a highly cited work in its area.

\subsection{Deduplication and screening process}

Before screening, record titles were normalised to reduce inconsistencies caused by punctuation, capitalisation, spacing, and character formatting.
During preprocessing, 27 records with blank titles could not be reliably processed for title-based comparison and were therefore excluded from deduplication.
The remaining 34,189 records were compared across all databases, resulting in the removal of 1,711 duplicate records and yielding 32,478 unique records for subsequent screening.

Screening was performed by a sequence of automated and manual stages.
First, 8 non-English records were removed, leaving 32,470 records.
Second, publication-format screening was applied.
Records explicitly identified as dissertations, preprints, editorials, news items, retracted items, or whole journal issues were excluded from the database-search workflow.
Reports identified through database metadata were also excluded, but were considered as support material when they satisfied the complementary source criteria described in Section~\ref{sec:complementary_sources}.
This distinction was used because reports, standards documents, ontology specifications, books, and thesis material provide ontology documentation and technical context, but do not form the main database-derived peer-reviewed study corpus. When a record was indexed differently across databases and at least one source classified it as an article, review, conference paper, workshop paper, or book chapter, the record was retained.
This step removed 3,916 records, leaving 28,554 records.

A scope-based pre-screening stage was then applied using the available title, abstract, and keyword metadata.
Records were retained when these metadata contained indications of both the building or energy domain and semantic modelling.
Building and energy domain indications included terms related to buildings, smart buildings, building systems, BEM/BEMS, HVAC, energy management, occupancy, thermal comfort, demand response, smart homes, smart grids, and building operation.
Semantic modelling indications included terms related to ontologies, semantic technologies, linked data, RDF, OWL, knowledge graphs, metadata, data models, information models, schemas, vocabularies, and taxonomies.
This stage removed 22,791 records, reducing the corpus to 5,763.

A rule-based title screening stage further decreased the record pool while avoiding premature exclusion of potentially relevant studies. At this stage, only the title was considered.
Records were retained when they contained at least one building or energy domain indication together with at least one indication from the semantic modelling, model-oriented, review-oriented, or application-oriented keyword groups.
Model-oriented terms included, for instance, model, modelling, framework, architecture, representation, interoperability, and integration. Review-oriented indications included terms such as review, survey, state of the art, comparison, and comparative. Application-oriented indications included terms such as optimisation, management, analytics, operation, monitoring, benchmarking, control, fault detection, diagnosis, and service.
This stage removed 3,893 records, leaving 1,870 records for manual abstract screening.

The abstracts of the remaining records were manually screened to identify records showing a clear ontology-related context and sufficient alignment with the survey scope.
This process resulted in 454 records for full-text and qualitative assessment.
After applying the inclusion and exclusion criteria described in Section~\ref{sec:inc_exc}, 68 journal, conference, and workshop papers were retained from the database-search workflow.
In addition, 1 technical report and 2 books were identified as ontology-relevant support sources, yielding a corpus of 71 sources from the PRISMA-oriented pipeline.

\enlargethispage{\baselineskip}
To support reproducibility, the scripts and supporting materials for the automated screening stages are publicly available in the project repository.\footnote{\url{https://github.com/Miracle-labmirx/prisma_oriented_screening_pipeline.git}} Appendix~\ref{app:screening_rules} summarises the keyword groups and screening logic.

\subsection{Inclusion and exclusion criteria}
\label{sec:inc_exc}
The qualitative assessment was designed to identify studies that made an explicit ontology-based contribution to building operational tasks or BEM applications. The inclusion criteria were:

\begin{enumerate}
    \item Studies representing building-related knowledge using graph-based semantic models in triple-based formats, such as RDF, OWL, or JSON-LD.
    \item Studies providing an explicit ontology-based semantic contribution, including the introduction of a new ontology, the extension of an existing ontology, or the integration of multiple semantic models within an application.
    \item Studies addressing building operational tasks or BEM applications, such as building energy monitoring, data integration, fault detection and diagnosis, control and optimisation, or building energy performance analysis.
\end{enumerate}

The exclusion criteria were:

\begin{enumerate}
    \item Studies whose knowledge representation did not rely on graph-based semantic models in triple-based formats, including studies based solely on non-graph-based representations such as relational schemas, XML, or UML.
    \item Studies focused primarily on building design, construction, or urban-scale analysis without a clear connection to building operations or BEM/BEMS.
    \item Studies that referred to semantics, metadata, interoperability, or ontologies only at a general level, without an explicit ontology-based contribution.
    \item Studies that did not address building operational tasks or BEM applications, such as monitoring, fault detection and diagnosis, control, optimisation, or energy analysis.
    \item Studies in which simulation, machine learning, control methods, sensor deployment, or visualisation formed the main contribution, without an explicit ontology-based semantic modelling.
    \item Studies conducted only at smart city, district, or community scale without a clear semantic representation of building-level operational knowledge.
    \item Studies for which the full text did not provide sufficient information to determine the ontology used, the operational application, or the semantic modelling contribution.
\end{enumerate}

\subsection{Complementary source identification}
\label{sec:complementary_sources}

The PRISMA-oriented selection resulted in 71 of the 150 ontology-relevant references.
The remaining 79 were included through complementary identification procedures.
These comprised additional journal, conference, and workshop papers as well as support materials such as ontology specifications, technical reports, online standards references, books, and thesis material.

The complementary identification process included backward reference checking, forward citation tracing, manual searches of key venues, targeted retrieval of seminal ontology frameworks, tracing of standards and project-related sources, and focused searches around relevant authors, standards bodies, and ontology framework names.

Complementary sources were included only when they documented an ontology or standard used in the reviewed studies, provided technical specifications needed to interpret an ontology, supplied methodological or domain background directly relevant to the analysed corpus, or represented a seminal ontology framework not adequately captured by the database search. They were not used to replace the database-derived peer-reviewed corpus, but to support interpretation, ontology documentation, and technical completeness.

The PRISMA-oriented selection and complementary identification procedures produced the final ontology-relevant set of 150 references, comprising 109 journal, conference, and workshop papers, 9 technical reports, 27 online ontology or standards references, 4 books, and 1 doctoral thesis.

\subsection{Data extraction and analysis procedure}
\label{sec:data_extraction_analysis}

The selected sources were analysed using a structured extraction and synthesis procedure aligned with the research questions. The purpose of this stage was to move from source identification to the characterisation of semantic models, their use in operational BEM applications, and the recurring modelling limitations observed across the reviewed literature.

Data extraction was performed manually from the full text and additional materials of each selected source. Where available, additional technical evidence was inspected, including official ontology documentation, ontology repositories, RDF/OWL serialisation files, appendices, semantic diagrams, implementation examples, and project specifications. This was necessary because ontology-based BEM studies often distribute semantic evidence across publications and associated technical artefacts.

The data extraction was organised into two complementary components. The first, supported the ontology review, where, for each ontology or semantic model, the extracted information included the ontology name, reference source, publication year, reused or extended ontologies, number of building-related classes, available serialisation formats, application scope, and main modelling focus. These data were used to construct the ontology catalogue and to support the characterisation of semantic models in Section~\ref{sec:semantic models}.
The second component supported the use case analysis. For each application-oriented study, the extracted information included the operational BEM task, the ontology or ontologies used, the building concepts represented, the ontology classes explicitly reported or traceable from the available evidence, the presence of newly created or extended concepts, and the modelling strategy adopted, including reuse, integration, specialisation, external inheritance, or application-specific extension. Concept-to-class traceability was recorded to show not only which ontologies were referenced, but also which ontology classes were used to represent the building concepts required by each operational BEM use case.

The synthesis was conducted in three steps. First, the extracted ontology-level information was synthesised qualitatively to describe the scope and modelling focus of the reviewed semantic models. Second, use case information was grouped according to the number of semantic models involved, distinguishing single-ontology, two-ontology, and multiple-ontology use cases. This grouping supported the analysis of how semantic models are applied in operational BEM and how semantic coverage gaps are addressed. 
Third, the extracted use case evidence was further analysed to assess semantic traceability and, where class-level evidence was available, quantitative coverage.

Semantic traceability was assessed using the OEC framework, while quantitative coverage was analysed using OIR and NTE. The detailed criteria, calculations, and use case findings are presented in Section~\ref{sec:usecases}. The extracted qualitative and quantitative patterns were finally synthesised to identify recurring limitations and future research needs, which are discussed in Section~\ref{sec:Discussion}.

\section{Context and background}
\label{sec:context_bkg}

This section addresses \emph{RQ1} and establishes the technical background for the survey.
It first distinguishes building information used during design and construction from the operational data required by BEM applications.
It then introduces the semantic concepts used to represent and connect building data. Finally, it identifies the core operational concepts that semantic models should capture for BEM.

\subsection{From BIM to operational building data}
\label{subsec:digitization}

The digitisation of building information is closely associated with the development of Building Information Modelling (BIM). Introduced in the 1970s, BIM has substantially influenced the \emph{Architecture, Engineering, and Construction} sector~\cite{shou2015comparative, eastman2011bim}. The National Building Information Model Standard defines BIM as a digital representation of the physical and functional characteristics of a facility and as a shared knowledge repository that supports decision-making throughout the facility life cycle~\cite{nbim}.

BIM is predominantly used during design and construction, where the emphasis is on geometrical and non-geometrical descriptions such as size, shape, volume, materials, costs, and schedules~\cite{eastman2011bim, pauwels2014supporting,borrmann2018building}. These data are valuable for planning and analysis, but they are not sufficient on their own for operational BEM, where applications must continuously process dynamic data from building systems, sensors, actuators, occupants, weather services, energy meters, and external energy resources.

The main open data model associated with BIM is the Industry Foundation Classes (IFC) standard~\cite{ifc}. IFC supports building-information exchange through several serialisations, including the STEP Physical File Format defined with EXPRESS~\cite{pratt2005iso}, XML-based representations~\cite{koronios2007integration}, and linked data representations such as ifcOWL~\cite{ifcOWL}. By making IFC information available in the Web Ontology Language (OWL), ifcOWL supports the publication and exchange of building information through semantic web technologies~\cite{beetz2005ontology, pauwels2016express, horrocks2004proposal}.

Despite its importance, IFC presents limitations for operational applications. BIM-to-IFC translation may produce incomplete or distorted models due to tool-specific export processes; schema adaptation is slow because it requires broad agreement across industries and software providers; and extension is difficult for many users because IFC is complex and historically tied to EXPRESS-based modelling~\cite{pauwels2017semantic}. These issues are especially relevant in the operational phase, where applications often integrate static and dynamic data from multiple sources.

The ifcOWL initiative advanced web-based building data exchange, but it also inherits much of IFC's complexity, making querying, extension, and modular reuse challenging in scalable building applications~\cite{terkaj2017method}. 
This motivated the development of more compact and modular Linked Building Data ontologies by the W3C Linked Building Data Community Group and related initiatives~\cite{pauwels2015linked,sarefcore, compton2012ssn,balaji2016brick}.
These models are discussed in Section~\ref{sec:semantic models}.

\subsection{Semantic web technologies for building data}
\label{subsec:advent}

Semantic models describe the meaning of data by defining concepts, properties, and relationships in a machine-interpretable form. They range from simple vocabularies and taxonomies to full ontologies: vocabularies identify terms, taxonomies add hierarchy, and ontologies provide richer graph-based representations in which entities are connected through explicitly defined relationships~\cite{w3c}. In information science, an ontology is commonly defined as an explicit specification of concepts, classes, objects, and relationships within a domain~\cite{gruber1993translation}, or as a ``formal, explicit specification of a shared conceptualisation''~\cite{studer1998knowledge}.

\begin{figure*}[!t]
\centering
\begin{minipage}{0.42\textwidth}
\subfigure[The Interconnection of domain-specific ontologies.]{
        \includegraphics[width=1\columnwidth]{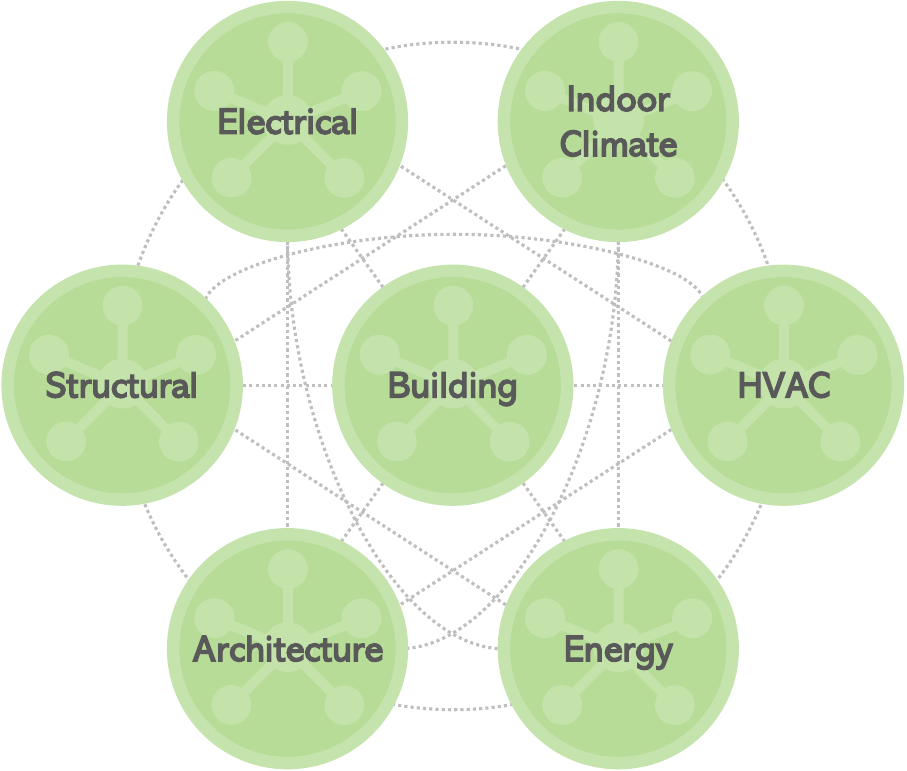}
        \label{fig:DSO}
}
\end{minipage}
\begin{minipage}{0.50\textwidth}
\centering
\subfigure[RDF Illustration.]{
	\includegraphics[width=0.75\columnwidth]{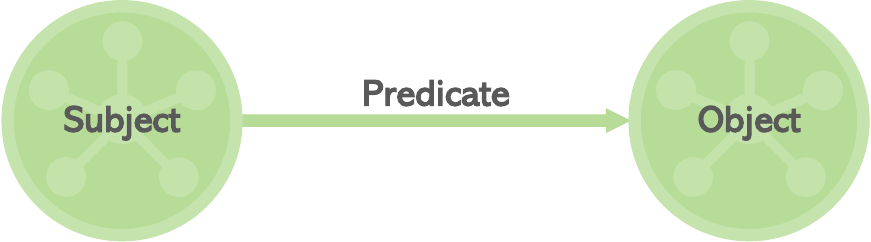}
	\label{fig:RDF}
}
\subfigure[NGSI-LD Information Model.]{
	\includegraphics[width=0.9\columnwidth]{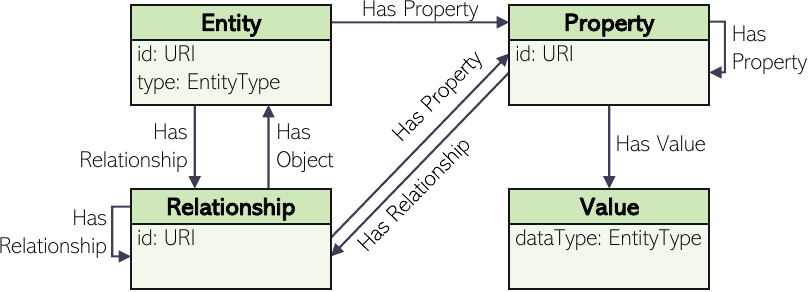}
	\label{fig:Ngsi-ld}
}
\end{minipage}

\caption{Key components of semantic data modelling and interoperability in smart environments.}
\label{fig:xxx2}
\end{figure*}

The semantic web extends the conventional web by enabling data representation in machine-readable and linkable formats~\cite{semanticweb}. Linked open data principles encourage publishing data using standard identifiers and graph-based representations to connect information from different sources~\cite{bizer2023linked}. The Resource Description Framework (RDF)~\cite{decker2000semantic} is central to this approach. It represents information as subject--predicate--object triples, forming graph patterns describing entities and their relationships, as illustrated in Figure~\ref{fig:RDF}. OWL builds on RDF to support richer ontology definitions, including class hierarchies, logical constraints, and reasoning.

The linked data approach allows connecting static building descriptions, device metadata, sensor observations, simulation data, time-series references, and external contextual data through shared identifiers and relationships. Figure~\ref{fig:DSO} illustrates this modular principle: domain-specific ontologies can remain separate while being linked where concepts overlap or complement one another. This modularity is important because no single ontology currently captures all concepts required by operational BEM applications.

New Generation Service Interface Linked Data (NGSI-LD)~\cite{privat2021guidelines} provides an information model for representing context entities and relationships in smart environments. It uses JSON-LD~\cite{sporny2020json} and identifies entities with Uniform Resource Locators or Uniform Resource Names~\cite{berners1994uniform, moats1997urn}, as shown in Figure~\ref{fig:Ngsi-ld}. Through context producers, context consumers, and a context broker, NGSI-LD manages and shares context information in real time and has been adopted in building applications to exchange operational context across heterogeneous systems~\cite{muatușa2025interoperable}.

For operational BEM, semantic web technologies are therefore not merely a data-formatting mechanism. They provide a layer for linking heterogeneous data sources, supporting reusable queries, enabling reasoning, and improving the portability of applications across buildings. The following section identifies the main building concepts that this semantic layer must represent.

\subsection{Identifying core concepts in building operational ontologies}
\label{sec:core concepts}

Concepts define the entity types represented in a semantic model, properties describe their attributes, and relationships specify how entities are associated~\cite{bergmann2020semantic}. In BEM, semantic models must represent both physical entities and operational abstractions. Core physical concepts include \emph{Building}, \emph{Zone}, \emph{Space}, \emph{Building Envelope}, \emph{Building System and Equipment}, \emph{Sensor}, \emph{Actuator}, \emph{Control Device}, \emph{Occupancy}, and external contextual data such as weather or energy price signals~\cite{pritoni2021metadata, gilani2020review}.

Spatial concepts are fundamental because most operational data are location dependent. A \texttt{space} denotes a tangible part of a building, such as a room, floor, corridor, or staircase, whereas a \texttt{zone} may correspond to a space, a group of spaces, or a virtual operational region used for control, monitoring, lighting, heating, cooling, or ventilation. This distinction matters because some ontologies model spaces and zones separately, whereas others represent them through broader classes.

The building envelope consists of elements that separate indoor and outdoor environments, such as walls, doors, roofs, and windows. These entities influence thermal behaviour, solar gains, air leakage, and energy demand. Although envelope concepts are often associated with design-stage data, they remain relevant during operation because they affect energy performance, comfort assessment, and simulation-based decision support.

Building systems and equipment include HVAC components, lighting systems, meters, renewable energy systems, storage devices, controllers, and other technical subsystems. Sensors observe properties such as temperature, humidity, illuminance, CO$_2$, occupancy, flow rate, and energy consumption, while actuators modify environmental or system states by controlling valves, dampers, fans, setpoints, windows, or lighting levels. These devices and relationships are central to automated monitoring, control, fault detection, and optimisation.

Occupancy is important in operational BEM because occupants influence both energy use and comfort. This survey distinguishes three related aspects: \emph{occupant behaviour}, which describes interactions with building systems such as opening windows, adjusting thermostats, or using appliances; \emph{occupant count}, which indicates the number of people in a space and affects internal loads and ventilation needs; and \emph{occupant detection}, which captures whether occupants are present, often inferred from sensors or smart devices. Existing ontologies represent these concepts at different levels of detail.

Energy flexibility and grid interaction also require semantic representation. Buildings increasingly interact with distributed energy resources, storage systems, renewable generation, electric vehicles, demand response programmes, and dynamic tariffs. Representing these entities and their relationships is necessary for coordinating buildings with smart grids and supporting decarbonisation objectives.

Operational BEM also requires semantic support for abstract concepts such as key performance indicators, assessments, control logic, services, computational workflows, schedules, and optimisation tasks, which connect raw operational data to decision-making. As discussed later in this survey, existing ontologies generally represent physical entities more consistently than computational and decision-support processes. This distinction is essential for evaluating the strengths and limitations of semantic models in intelligent context-aware BEM systems.

For operational BEM applications, ontologies provide a shared semantic structure, needed to integrate heterogeneous and continuously changing building data, its technical systems, occupants, and surrounding context. With respect to \emph{RQ1}, they transform fragmented operational data into interpretable, linkable, and reusable explicit representations of building context, system behaviour, and decision-support requirements, supporting more portable BEM applications.

\section{Semantic models}
\label{sec:semantic models}

This section identifies the semantic models that have been used to represent building operational knowledge, addressing the research question \emph{RQ2}. Its aim is not to compare these models as competing alternatives, but to clarify their modelling scope, the building operational concepts they support, and the roles they play in BEM-related applications. The section first discusses core semantic models that are frequently used in building operation, before reviewing additional domain-specific and supporting ontologies that extend semantic representation towards smart homes, automation, energy flexibility, smart grids, and performance assessment.

Table~\ref{tab:other-ontology} provides the detailed catalogue of the sixty-one ontologies reviewed for the building operational phase, including references, publication years, reused or extended ontologies, class counts, serialisation formats, application scopes, and main focus areas. Class counts were taken from official documentation when available; otherwise, Protégé\footnote{\url{https://protege.stanford.edu/}} and ontospy\footnote{\url{https://lambdamusic.github.io/Ontospy/}} were used to extract class-count statistics. The scope icons in Table~\ref{tab:other-ontology} distinguish smart building, smart home, smart city, smart device, smart energy, and smart grid applications, following established definitions of these domains~\cite{buckman2014smart,marikyan2019systematic,camero2019smart,lund2014renewable,tuballa2016review,lazar2015we}.

\subsection{Core semantic models}
\label{sec:core semantic models}

This section reviews most relevant (core) models to building operations. ifcOWL is included as a foundational BIM linked-data model, while BOT, SAREF, SSN/SOSA, and Brick are discussed in greater detail because they are widely adopted in practical use cases. Project Haystack~\cite{projecthaystack} is included because its tag-based representation influenced Brick. XML and UML data models are excluded because they are less aligned with linked data methods and W3C semantic web principles~\cite{decker2000semantic, hitzler2009owl}.

\subsubsection{\colorbox{ifc}{ifcOWL}---Industry foundation classes in OWL}
\label{sec:ifcowl}

ifcOWL\phantomsection\label{ont:ifcowl}~\cite{ifcOWL} is the OWL representation of the IFC schema. It makes IFC building-model information available to semantic web technologies that support linking, querying, and reasoning over RDF/OWL data.

\begin{longtblr}[
  caption = {An overview of the analysed key ontologies.},
  label = {tab:other-ontology},
  remark{{\footnotesize Table Note}} = {{\footnotesize In this table, the ontology names are linked to where they are explained in more detail in the text and the numbers of classes are linked to the repositories where the ontologies are publicly available on the Internet. Shaded group rows identify the ontology categories discussed in Section~\ref{sec:semantic models}; thicker horizontal rules mark the boundaries between these groups. \newline
  \footnotemark[1]{\cite{realEstateCore}}, \footnotemark[2]{\cite{salameh2019ssg}}, \footnotemark[3]{\cite{efficient}}, \footnotemark[4]{\cite{blanco2021comparison}}}},
]{
  colspec = {X[0.75]X[0.75]X[0.5]X[1.05]X[0.65]X[0.95]X[0.5]X[1.82]},
  cells   = {font = \footnotesize\selectfont},
  rowhead = 1,
  rowfoot = 2,
  hlines,
  stretch=0,
  rows={ht=\baselineskip},
}
\cline[2pt]{-}
\textbf{Ontology} & \textbf{Ref.} & \textbf{Year} & \textbf{Uses/\newline Extends} & \textbf{Class\-es} & \textbf{Formats} & \textbf{Scope}   & \textbf{Main Focus}  \\
\cline[2pt]{-}
\SetCell[c=8]{c, bg=s4b, font=\footnotesize\bfseries} Core semantic models & & & & & & & \\
\cline[2pt]{-}
\hyperref[ont:ifcowl]{\colorbox{ifc}{ifcOWL}} & \cite{pauwels2016express} & 2016 & None & \href{https://github.com/buildingsmart-community/ifcOWL/blob/d450645d/IFC4_ADD2.ttl}{1286} &\ftag{O}\ftag{R}\ftag{T} & \includegraphics[height=1.3em] {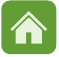}\includegraphics[height=1.3em]{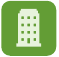} & IFC-based BIM linked data\\
\hyperref[ont:bot]{\colorbox{bot}{BOT}} & \cite{botOntology} & 2017 & None &  \href{https://github.com/w3c-lbd-cg/bot}{8} & \ftag{T} & \includegraphics[height=1.3em] {smart_home.pdf}\includegraphics[height=1.3em]{smart_building.pdf}& Bldg. topology\\

\hyperref[ont:saref]{\colorbox{sar}{SAREF}} & \cite{sarefcore} & 2015 & None & \href{https://labs.etsi.org/rep/saref/saref-core}{29} & \ftag{$\ast$}& \includegraphics[height=1.3em] {smart_home.pdf}\includegraphics[height=1.3em]{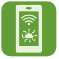} & Energy efficiency\\

\hyperref[ont:ssnsosa]{\colorbox{ssn}{SSN}/ \colorbox{sos}{SOSA}} & \cite{compton2012ssn, janowicz2019sosa} & 2019 & None & \href{https://github.com/w3c/sdw-sosa-ssn}{21} & \ftag{T} & \includegraphics[height=1.3em]{smart_devices.pdf} & Sensors, measurements\\

\hyperref[ont:brick]{\colorbox{bri}{Brick}} & \cite{balaji2016brick, luo2022extending} & 2016 & \colorbox{oth}{PH} & \href{https://github.com/BrickSchema/Brick}{1450} & \ftag{T} &
\includegraphics[height=1.3em]{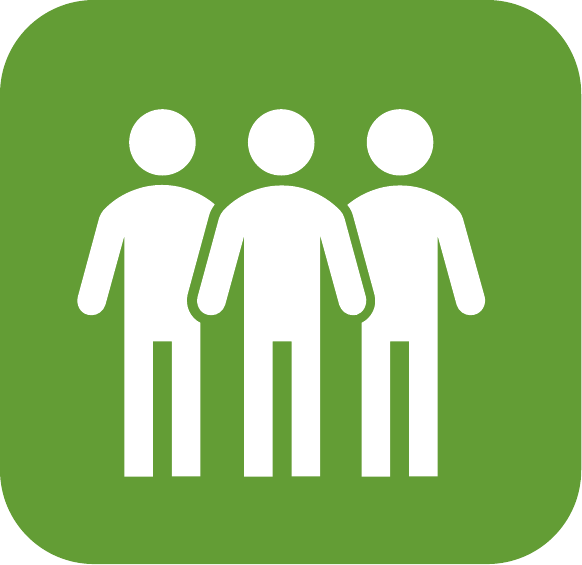}
\includegraphics[height=1.3em]{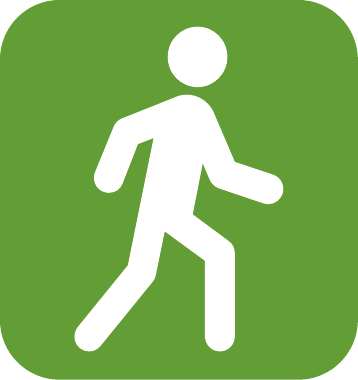}
\includegraphics[height=1.3em]{smart_devices.pdf}\includegraphics[height=1.3em] {smart_home.pdf}\includegraphics[height=1.3em] {smart_building.pdf}& Occupancy, behaviour and Bldg. entities\\

\hyperref[ont:ph]{\colorbox{oth}{PH}} & \cite{projecthaystack}  & 2014 & None & \href{https://github.com/Project-Haystack}{514*} & \ftag{T}\ftag{J} & \includegraphics[height=1.3em]{smart_devices.pdf}\includegraphics[height=1.3em] {smart_home.pdf}\includegraphics[height=1.3em] {smart_building.pdf} & Bldg. info. from devices\\
\cline[2pt]{-}
\SetCell[c=8]{c, bg=s4b, font=\footnotesize\bfseries} Smart home, device, occupancy, and ambient-intelligence ontologies & & & & & & & \\
\cline[2pt]{-}
\hyperref[ont:thinkhome]{ThinkHome} & \cite{reinisch2010thinkhome}  & 2010 & None & \href{https://www.auto.tuwien.ac.at/downloads/thinkhome/ontology/}{1152} & \ftag{O} &\includegraphics[height=1.3em] {smart_home.pdf} & Energy for homes\\

\hyperref[ont:dogont]{DogOnt} & \cite{bonino2008dogont} & 2008 & None & \href{https://iot-ontologies.github.io/dogont/documentation/index-en.html}{1110} & \ftag{$\ast$} & \includegraphics[height=1.3em] {smart_building.pdf} & Devices  \\

\hyperref[ont:poweront]{PowerOnt} & \cite{bonino2015poweront} & 2015 & DogOnt & \href{https://github.com/iot-ontologies/poweront/blob/master/poweront.owl}{1070} & \ftag{O} & \includegraphics[height=1.3em]{smart_home.pdf}\includegraphics[height=1.3em]{smart_devices.pdf} & Energy efficiency\\

\hyperref[ont:homeapplianceontology]{Home~Appl.} & \cite{shah2011ontology} & 2011 & None & N/A & \rtag{C}\ftag{O} &
\includegraphics[height=1.3em]{smart_home.pdf}\includegraphics[height=1.3em]{smart_building.pdf} & Energy consumption \\

\hyperref[ont:bonsai]{BonSAI} & \cite{stavropoulos2012bonsai} & 2012 & None & \href{ https://github.com/BONSAMURAIS/BONSAI-ontology-RDF-framework}{84} &  \ftag{R}\ftag{X} & \includegraphics[height=1.3em] {smart_building.pdf} & Ambient intelligence \\

\hyperref[ont:oncom]{ONCOM} & \cite{orozco2019ontology} & 2019 & \colorbox{ssn}{SSN}/\colorbox{sos}{SOSA}  & N/A & \rtag{C}\ftag{O} &
\includegraphics[height=1.3em]{smart_home.pdf}\includegraphics[height=1.3em]{smart_building.pdf} & Thermal Comfort \\
\hyperref[ont:aalontology]{AAL ontology} & \cite{marcello2024digital} & 2024 & None & 24 & \rtag{C} & \includegraphics[height=1.3em]{smart_home.pdf}
\includegraphics[height=1.3em]{smart_building.pdf} \includegraphics[height=1.3em]{smart_devices.pdf} \includegraphics[height=1.3em]{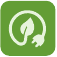} & Human-building interaction
\\
\hyperref[ont:op]{OP} & \cite{opontology} & 2020 & \colorbox{sar}{SAREF}  & \href{https://bimerr.iot.linkeddata.es/def/occupancy-profile/}{66} & \ftag{$\ast$} &
\includegraphics[height=1.3em]{OB.pdf}
\includegraphics[height=1.3em]{smart_home.pdf}\includegraphics[height=1.3em]{smart_building.pdf} & Occupant behaviour and Bldg. energy profiles\\
\cline[2pt]{-}
\SetCell[c=8]{c, bg=s4b, font=\footnotesize\bfseries} Building automation, control, and technical-system ontologies & & & & & & & \\
\cline[2pt]{-}

\hyperref[ont:realestatecore]{RealEstate\-Core} & \cite{realEstateCore}  & 2019 & \colorbox{ssn}{SSN}/\colorbox{sos}{SOSA} & \href{https://github.com/RealEstateCore/rec}{178}\footnotemark[1] & \ftag{R} & \includegraphics[height=1.3em] {smart_building.pdf}\includegraphics[height=1.3em] {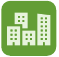} & Bldg. in cities \\

\hyperref[ont:bacs]{BACS} & \cite{terkaj2017reusing} & 2017 & \colorbox{bot}{BOT} \colorbox{ssn}{SSN}/\colorbox{sos}{SOSA} & \href{https://github.com/difactory/repository/tree/main/ontoeng}{9} & \ftag{O} & \includegraphics[height=1.3em] {smart_building.pdf} & Automation and control \\

\hyperref[ont:basont]{BASont} & \cite{ploennigs2012basont} & 2012 & None & N/A & \rtag{C} & \includegraphics[height=1.3em] {smart_building.pdf} & Automation systems \\

\hyperref[ont:sbms]{SBMS} & \cite{kuvcera2018semantic}  & 2018 & \colorbox{ssn}{SSN} & \href{https://is.muni.cz/www/akucera/sbms/v1_0/SemanticBMS.owl}{25} & \ftag{O} &
\includegraphics[height=1.3em]{smart_building.pdf}\includegraphics[height=1.3em]{smart_devices.pdf} & Bldg. automation systems \\

\hyperref[ont:tso]{TSO} & \cite{pauen2024integrated} & 2024 & None & 40 & \ftag{T}\ftag{R}\ftag{J} & \includegraphics[height=1.3em] {smart_building.pdf} & Building technical systems \\

\hyperref[ont:fso]{FSO} & \cite{kukkonen2022ontology} & 2022 & None & \href{https://alikucukavci.github.io/FSO/}{14} & \ftag{T} & \includegraphics[height=1.3em] {smart_building.pdf} & Mass/energy flow \\

\hyperref[ont:fpo]{FPO} & \cite{kucukavci2022taking} & 2023 & \colorbox{oth}{FSO} & 50\footnotemark[3]
& \rtag{C} & \includegraphics[height=1.3em] {smart_building.pdf} & HVAC components \\

\hyperref[ont:ctrlont]{CTRLont} & \cite{schneider2017ontology} & 2019 & None & \href{https://technicalbuildingsystems.github.io/Ontologies/CTRLont/index-en.html}{11} & \ftag{$\ast$} &
\includegraphics[height=1.3em] {smart_building.pdf}
\includegraphics[height=1.3em] {smart_energy.pdf}
\includegraphics[height=1.3em] {smart_devices.pdf} & Building control applications
\\
\hyperref[ont:ashrae]{ASHRAE 223p} & \cite{ashrae223p} & 2023 & None & N/A & \rtag{C} &
\includegraphics[height=1.3em]{smart_home.pdf}
\includegraphics[height=1.3em] {smart_building.pdf}
\includegraphics[height=1.3em] {smart_devices.pdf}
\includegraphics[height=1.3em] {smart_energy.pdf} & Building analytics and Automation

\\
\hyperref[ont:google digital buildings]{Digital \newline Buildings} & \cite{google, berkoben2020digital} & 2020 & \colorbox{bri}{Brick} \colorbox{oth}{PH} & \href{ https://github.com/google/digitalbuildings}{1308} & \ftag{R} & \includegraphics[height=1.3em]{smart_building.pdf} & Bldg. equipment\\

\hyperref[ont:vbschema]{VB schema} & \cite{lee2025metadata} & 2025 & None & \href{https://skku-bist.github.io/introduction}{3} & \ftag{R} &
\includegraphics[height=1.3em]{smart_home.pdf}
\includegraphics[height=1.3em]{smart_building.pdf} & Ontology-enabled AI-driven BEM \\

\hyperref[ont:sbonto]{SBonto} & \cite{vzavcek2017sbonto} & 2017 & None & N/A & \rtag{C} & \includegraphics[height=1.3em]{smart_building.pdf} & Smart bldg. \\

\hyperref[ont:ontosb]{Onto-SB} & \cite{degha2019intelligent} & 2019 & None & N/A & \rtag{C}\ftag{O}& \includegraphics[height=1.3em]{smart_building.pdf} & Smart bldg. \\
\cline[2pt]{-}
\SetCell[c=8]{c, bg=s4b, font=\footnotesize\bfseries} Energy flexibility, demand response, and grid-interactive ontologies & & & & & & & \\
\cline[2pt]{-}

\hyperref[ont:efont]{EFOnt} & \cite{li2022semantic}  & 2022 & None & \href{https://github.com/LBNL-ETA/EnergyFlexibilityOntology}{87} & \ftag{$\ast$} & \includegraphics[height=1.3em] {smart_building.pdf}\includegraphics[height=1.3em] {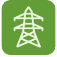} & Energy flexibility\\

\hyperref[ont:mirabel]{Mirabel} & \cite{verhoosel2012ontology} & 2012 & None & 5\footnotemark[2] & \rtag{C} &
\includegraphics[height=1.3em] {smart_devices.pdf} & Energy flexibility \\

\hyperref[ont:sargon]{SARGON} & \cite{haghgoo2020sargon} & 2020 & \colorbox{sar}{SAREF} & \href{https://github.com/N5GEH/n5geh.datamodel}{169} & \ftag{T}\ftag{O} & \includegraphics[height=1.3em] {smart_grid.pdf}\includegraphics[height=1.3em] {smart_building.pdf} & Bldg. entities  \\

\hyperref[ont:sargon2]{SARGON2} & \cite{tun2025sargon2} & 2025 & SARGON EM-KPI & \href{https://sumon-tun.github.io/SARGON2/OnToology/sargon2.owl/documentation/index-en.html}{2109} & \ftag{$\ast$} &
\includegraphics[height=1.3em] {smart_building.pdf}
\includegraphics[height=1.3em] {smart_grid.pdf}
\includegraphics[height=1.3em] {smart_energy.pdf}  & Building energy systems integration
\\

\hyperref[ont:respond]{RESPOND} & \cite{respond} & 2020 & \colorbox{bot}{BOT} \colorbox{oth}{SEAS} \colorbox{sar}{SAREF} & \href{https://respond-project.github.io/RESPOND-Ontology/respond/index-en.html}{45} & \ftag{$\ast$} &
\includegraphics[height=1.3em]{smart_building.pdf}\includegraphics[height=1.3em] {smart_grid.pdf} & Energy dispatching\\

\hyperref[ont:openadr]{OpenADR} & \cite{fernandez2020openadr} & 2020 & None & \href{https://albaizq.github.io/OpenADRontology/OnToology/ontology/openADRontology.owl/documentation/index-en.html}{76} & \ftag{$\ast$} &
\includegraphics[height=1.3em]{smart_grid.pdf}\includegraphics[height=1.3em]{smart_building.pdf} & Demand-response \\

\hyperref[ont:delta]{DELTA} & \cite{fernandez2021supporting} & 2021 & OpenADR
\newline\colorbox{sar}{SAREF} & \href{https://github.com/oeg-upm/delta-ontology}{59} & \ftag{$\ast$} & \includegraphics[height=1.3em]{smart_grid.pdf}\includegraphics[height=1.3em]{smart_building.pdf} &  DR energy market\\

\hyperref[ont:sg-bems]{SG-BEMS} & \cite{schachinger2016ontology} & 2016 & None & N/A & \rtag{C}\ftag{O}\ftag{T} &
\includegraphics[height=1.3em]{smart_grid.pdf}\includegraphics[height=1.3em]{smart_building.pdf} &  BEMS \\

\hyperref[ont:dedalus]{DEDALUS} & \cite{dedalus} & 2024 & dCO \colorbox{sar}{SAREF} \colorbox{bri}{Brick} NGSI-LD & \href{https://engsep.github.io/dedalus-ontology/index-en.html}{220} & \ftag{$\ast$} &
\includegraphics[height=1.3em]{smart_building.pdf} \includegraphics[height=1.3em]{smart_energy.pdf}  \includegraphics[height=1.3em]{smart_devices.pdf} & Demand response services\\
\hyperref[ont:dco]{dCO} & \cite{dco} & 2022 & None & \href{https://www.dco.domos-project.eu/}{241} & \ftag{$\ast$} &
\includegraphics[height=1.3em]{smart_building.pdf} \includegraphics[height=1.3em]{smart_energy.pdf}  \includegraphics[height=1.3em]{smart_devices.pdf} & Demand response services
\\
\cline[2pt]{-}
\SetCell[c=8]{c, bg=s4b, font=\footnotesize\bfseries} Energy-system, smart-grid, and cross-domain interoperability ontologies & & & & & & & \\
\cline[2pt]{-}
\hyperref[ont:seas]{SEAS} & \cite{lefranccois2017seas} & 2017 & None & \href{https://github.com/thesmartenergy/seas}{293} & \ftag{$\ast$} & \includegraphics[height=1.3em]{smart_building.pdf} & Energy analysis \\

\hyperref[ont:oeo]{OEO} & \cite{booshehri2021introducing} & 2021 & None & \href{https://github.com/OpenEnergyPlatform/ontology}{928} & \ftag{O} & \includegraphics[height=1.3em]{smart_energy.pdf} & Energy system analysis\\

\hyperref[ont:sg1]{SG\textsuperscript{1}} & \cite{gillani2014generic} & 2014 & None & N/A & \rtag{C}\ftag{T} &  \includegraphics[height=1.3em]{smart_grid.pdf} & Real-time management \\

\hyperref[ont:sg2]{SG\textsuperscript{2}} & \cite{tefek2023smart} & 2023 & None & \href{https://github.com/smartgridadsc/Ontology_CVE_Populator/blob/main/Paper_ontology.owl}{75} & \ftag{O} &
\includegraphics[height=1.3em]{smart_grid.pdf} & Vulnerability assessments \\

\hyperref[ont:ssg]{SSG} & \cite{salameh2019ssg} & 2019 & None & N/A & \rtag{C}\ftag{O}\ftag{T}&  \includegraphics[height=1.3em] {smart_grid.pdf} & Smart grid component\\

\hyperref[ont:ssgim]{SSGIM} & \cite{zhou2012semantic} & 2012 & None & N/A & \rtag{C}\ftag{O}&  \includegraphics[height=1.3em] {smart_grid.pdf} & DR applications\\

\hyperref[ont:sepa's sg]{SEPA's SG} & \cite{sepa} & 2020 & None & \href{https://github.com/smart-electric-power-alliance/Electric-Grid-Ontology}{250}\footnotemark[4] & \ftag{T} & \includegraphics[height=1.3em]{smart_grid.pdf} & Smart grid applications \\

\hyperref[ont:oema]{OEMA} & \cite{cuenca2017unified} & 2017 & ThinkHome, SAREF4ENER, and SG\textsuperscript{1} & \href{https://innoweb.mondragon.edu/ontologies/oema/ontologynetwork/1.1/index-en.html}{24} & \ftag{$\ast$}& \includegraphics[height=1.3em]{smart_grid.pdf}\includegraphics[height=1.3em]{smart_energy.pdf} & Different energy domains \\

\hyperref[ont:dabgeo]{DABGEO} & \cite{cuenca2020dabgeo} & 2020 & OEMA & \href{https://innoweb.mondragon.edu/ontologies/dabgeo/index-en.html}{1965} & \ftag{O}& \includegraphics[height=1.3em]{smart_grid.pdf}\includegraphics[height=1.3em]{smart_energy.pdf} & Different energy domains \\

\hyperref[ont:cim]{CIM} & \cite{cim} & 2015 & None & \href{ http://cim.puffinsemantics.com/}{39} & \ftag{O} & \includegraphics[height=1.3em]{smart_grid.pdf} &  Model profile \\

\hyperref[ont:newoseim]{(New) \newline OSEIM}& \cite{saba2019development,saba2021ontology}& 2019 & None & N/A & \rtag{C}\ftag{O} & \includegraphics[height=1.3em]{smart_grid.pdf}\includegraphics[height=1.3em]{smart_building.pdf}&Electrical energy consump.\\
\cline[2pt]{-}
\SetCell[c=8]{c, bg=s4b, font=\footnotesize\bfseries} Performance assessment, KPI, simulation, and building energy analysis ontologies & & & & & & & \\
\cline[2pt]{-}

\hyperref[ont:eeont]{EEOnt} & \cite{diaz2013eeont} & 2013 & None & N/A & \rtag{C} &
\includegraphics[height=1.3em] {smart_building.pdf} & Energy Efficiency \\

\hyperref[ont:eepsa]{EEPSA} & \cite{esnaola2021eepsa} & 2021 & None & 258 & \ftag{$\ast$} & \includegraphics[height=1.3em]{smart_building.pdf} & Smart bldg. \\

\hyperref[ont:bem]{\textit{Lork et al.’s} \newline ontology} & \cite{lork2019ontology} & 2019 & \colorbox{ssn}{SSN} & N/A & \rtag{C}\ftag{O} & \includegraphics[height=1.3em]{smart_building.pdf}\includegraphics[height=1.3em]{smart_devices.pdf} &  Energy optimisation\\

\hyperref[ont:em-kpi]{EM-KPI} & \cite{li2019enhancing} & 2019 & \colorbox{ssn}{SSN} ThinkHome & \href{http://energy.linkeddata.es/em-kpi/ontology/index-en.html}{133} & \ftag{$\ast$} & \includegraphics[height=1.3em]{smart_building.pdf} & KPI for bldg. perf. \\

\hyperref[ont:kpi]{KPI} & \cite{kpiOntology} & 2020 & \colorbox{oth}{saref4city} \colorbox{sar}{SAREF}& \href{https://bimerr.iot.linkeddata.es/def/key-performance-indicator/}{17} & \ftag{$\ast$} & \includegraphics[height=1.3em]{smart_building.pdf}\includegraphics[height=1.3em]{smart_city.pdf} & KPI for bldg. renovation  \\

\hyperref[ont:pf]{PF} & \cite{corry2015performance} & 2015 & \colorbox{ssn}{SSN} & N/A & \rtag{C} & \includegraphics[height=1.3em]{smart_building.pdf} & Bldg. perf. analysis\\

\hyperref[ont:bop]{BOP} & \cite{donkersbuilding} & 2021 & None & \href{https://alexdonkers.github.io/bop/}{18} & \ftag{$\ast$} &
\includegraphics[height=1.3em]{smart_building.pdf} & Bldg. perf. analysis\\

\hyperref[ont:semanco]{SEMANCO} & \cite{madrazo2012semanco}  & 2012 & None & \href{http://semanco-tools.eu/ontology-releases/eu/semanco/ontology/SEMANCO/SEMANCO.owl?}{980} & \ftag{O} & \includegraphics[height=1.3em]{smart_city.pdf}\includegraphics[height=1.3em]{smart_energy.pdf} & Urban plan., energy mgmt.\\

 \hyperref[ont:ero]{EPG-RISK ontology} & \cite{yilmaz2026developing} & 2026 & None & 3 & \ftag{O} &  \includegraphics[height=1.3em]{smart_building.pdf} \includegraphics[height=1.3em]{smart_energy.pdf}  \includegraphics[height=1.3em]{smart_devices.pdf} & Building Energy Performance\\
\hyperref[ont:woso]{WoSO} & \cite{hounas2024web} & 2025 & \colorbox{sar}{SAREF} & \href{https://zhounas.github.io/woso/index-en.html}{6} & \ftag{$\ast$} &
\includegraphics[height=1.3em]{smart_building.pdf} \includegraphics[height=1.3em]{smart_energy.pdf}  \includegraphics[height=1.3em]{smart_devices.pdf} & Building performance simulation
\\

\hyperref[ont:sesame]{SESAME} & \cite{fensel2013sesame} & 2010 & None & N/A & \rtag{C}\ftag{O}\ftag{N} &
\includegraphics[height=1.3em]{smart_grid.pdf}\includegraphics[height=1.3em]{smart_building.pdf} & Energy Optimisation \\

\hyperref[ont:bigg]{BIGG} & \cite{broto2025interoperable} & 2025 & \colorbox{sar}{SAREF} \colorbox{ssn}{SSN}& \href{https://github.com/BeeGroup-cimne/biggontology}{56} & \ftag{T} & \includegraphics[height=1.3em] {smart_building.pdf} \includegraphics[height=1.3em] {smart_city.pdf} \includegraphics[height=1.3em] {smart_energy.pdf} \includegraphics[height=1.3em] {smart_devices.pdf} & Building energy analysis \\
\hyperref[ont:icbms]{ICBMS} & \cite{ji2017intelligent} & 2017 & \colorbox{sar}{SAREF} & N/A & \rtag{C} & \includegraphics[height=1.3em]{smart_building.pdf} & Thermal comfort analysis\\

\emph{Scopes}: & \SetCell[c=7]{}Smart \includegraphics[height=1.0em]{smart_devices.pdf} Device, \includegraphics[height=1.0em]{smart_home.pdf} Home, \includegraphics[height=1.0em]{smart_building.pdf} Building, \includegraphics[height=1.0em]{smart_city.pdf} City, \includegraphics[height=1.0em]{smart_energy.pdf} Energy, \includegraphics[height=1.0em]{smart_grid.pdf} Grid,
\includegraphics[height=1.0em]{OccupantNumber.pdf} Occ.,
\includegraphics[height=1.0em]{OB.pdf} Behav.,
\includegraphics[height=1.0em]{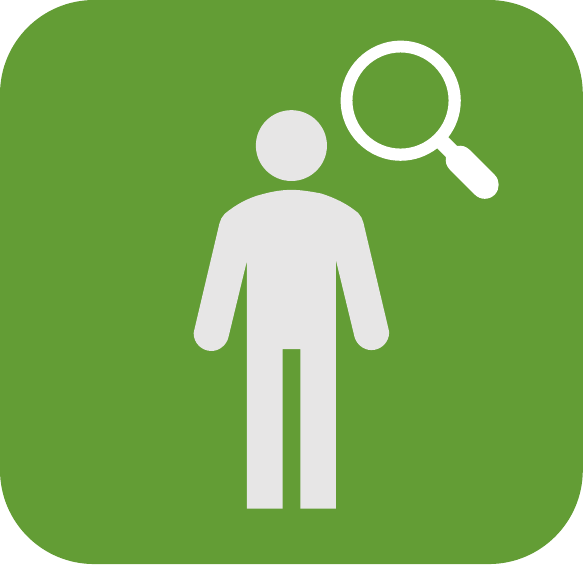} Detect.\\

\cline[white]{-}

\emph{Formats}: & \SetCell[c=7]{}\ftag{T}Turtle, \ftag{R}RDF, \ftag{X}XML, \ftag{O}OWL, \ftag{J}JSON-LD, \ftag{N}N-Triples, \ftag{$\ast$}Many, \rtag{C}Closed Source\\

\cline[white]{-}
\end{longtblr}

\noindent By exposing BIM information as linked data, ifcOWL enables IFC-based building descriptions to be connected with other RDF/OWL resources~\cite{beetz2005ontology,pauwels2016express,horrocks2004proposal}. For operational BEM, it mainly provides detailed design and asset information that can contextualise integration with dynamic data sources.

However, ifcOWL also inherits much of IFC's breadth and complexity, which can make lightweight and modular use in operational BEM studies challenging~\cite{terkaj2017method,petrova2021meaning}.
This follows from its direct translation from the IFC schema, which keeps close alignment with the IFC standard and retains its original modelling structure. In this survey, ifcOWL is therefore treated as a foundational BIM linked-data model and supporting ontology, while the core operational analysis focuses on semantic models more directly used in building operation use cases.

\subsubsection{\colorbox{bot}{BOT}---Building topology ontology}
\label{sec:bot}

BOT\phantomsection\label{ont:bot}~\cite{rasmussen2021bot} provides a lightweight and extensible model for representing building topology, including sites, buildings, storeys, spaces, elements, interfaces, and 3D geometry. Developed by the Linked Building Data Community Group under the W3C Consortium, BOT uses linked data and semantic web technologies to support web-based data integration across AECO applications.

\begin{figure}[ht!]
    \centering
    \includegraphics[width=\linewidth]{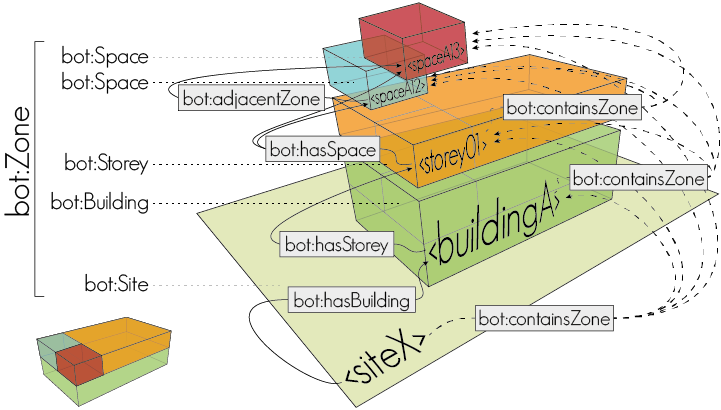}
    \caption{Representation of Building Topology by BOT (image from~\cite{botOntology}).}
    \label{fig:BOT}
\end{figure}

BOT focuses on the structural and spatial organisation of buildings, as shown in Figure~\ref{fig:BOT}. Because it is intentionally lightweight, it is often combined with other ontologies to represent product details, sensor observations, IoT devices, complex geometry, or project management data. Its main classes include \texttt{Zone}, \texttt{Site}, \texttt{Building}, \texttt{Storey}, \texttt{Space}, \texttt{Element}, and \texttt{Interface}.

\subsubsection{\colorbox{sar}{SAREF}---Smart applications REFerence ontology and its extensions}
\label{sec:saref}

SAREF\phantomsection\label{ont:saref} is listed in Table~\ref{tab:other-ontology} as a reference ontology for IoT applications~\cite{sarefcore}. Developed by the European Commission and ETSI with input from the smart-appliances industry~\cite{daniele2015created}, it addresses fragmentation across IoT standards, platforms, and technologies by providing a shared model for device communication~\cite{etsi2016103, etsi2016103b, daniele2018study}. ETSI publishes SAREF as a core ontology with domain-specific extensions covering energy, environment, buildings, smart cities, industry, agriculture, automotive systems, eHealth, wearables, water, lifts, smart grids, maritime applications, and systems~\cite{saref4ener,saref4envi,saref4bldg,saref4city,saref4inma,saref4agri,saref4auto,saref4ehaw,saref4wear,saref4watr,saref4lift,saref4grid,saref4mari}. A preliminary SAREF-based solution was demonstrated in commercial energy-domain products in 2017~\cite{moreira2017towards}. In this review, the discussion focuses on SAREF4BLDG, SAREF4ENER, and SAREF4SYST because of their relevance to buildings, while SAREF4GRID is noted for grid-interactive applications. Table~\ref{tab:my-table1} reports the class, property, and individual counts for SAREF and its extensions.

\begin{table}[b!]
\centering
\caption{SAREF and its extension's ontological specification}
\label{tab:my-table1}
\scriptsize
\setlength{\tabcolsep}{4pt}
\renewcommand{\arraystretch}{1.12}
\begin{tabular*}{\columnwidth}{@{\extracolsep{\fill}}lrccc@{}}
\toprule
Ontology & Cls. & Obj. & Data & Indiv. \\
\midrule
\colorbox{sar}{SAREF}       & 29      & 62                & 8               & 0                 \\
\hline
\colorbox{s4s}{SAREF4SYST}  & 3       & 9                 & 0               & 0                 \\
\hline
SAREF4ENER  & 76      & 96                & 75              & 50                \\
\hline
\colorbox{s4b}{SAREF4BLDG}  & 65      & 6                 & 81              & 0                 \\
\hline
SAREF4ENVI  & 66      & 25                & 6               & 62                \\
\hline
SAREF4CITY  & 14      & 15                & 6               & 0                 \\
\hline
SAREF4INMA  & 21      & 16                & 10              & 0                 \\
\hline
SAREF4AGRI  & 21      & 8                 & 4               & 10                \\
\hline
SAREF4AUTO  & 48      & 5                 & 7               & 584               \\
\hline
SAREF4EHAW  & 21      & 12                 & 10              & 12                \\
\hline
SAREF4WEAR  & 29      & 19                & 6               & 12                \\
\hline
SAREF4WATR  & 49      & 8                 & 11              & 76               \\
\hline
SAREF4LIFT & 25      & 12                & 15              & 5 \\
\hline
SAREF4GRID & 40      & 25                & 42             & 28 \\
\hline
SAREF4MARI & 62      & 18                & 50             & 177 \\
\hline
\end{tabular*}
\vspace{2pt}
\begin{minipage}{\columnwidth}
\scriptsize\emph{Table note:} Cls. = classes; Obj. = object properties; Data = data properties; Indiv. = named individuals.
\end{minipage}
\end{table}

The counts in Table~\ref{tab:my-table1} include only classes, object properties, data properties, and named individuals declared in the namespace of each listed ontology or extension; imported or reused terms from external namespaces are excluded.

 \begin{figure*}[t!]
    \centering
    \includegraphics[width=1\linewidth]{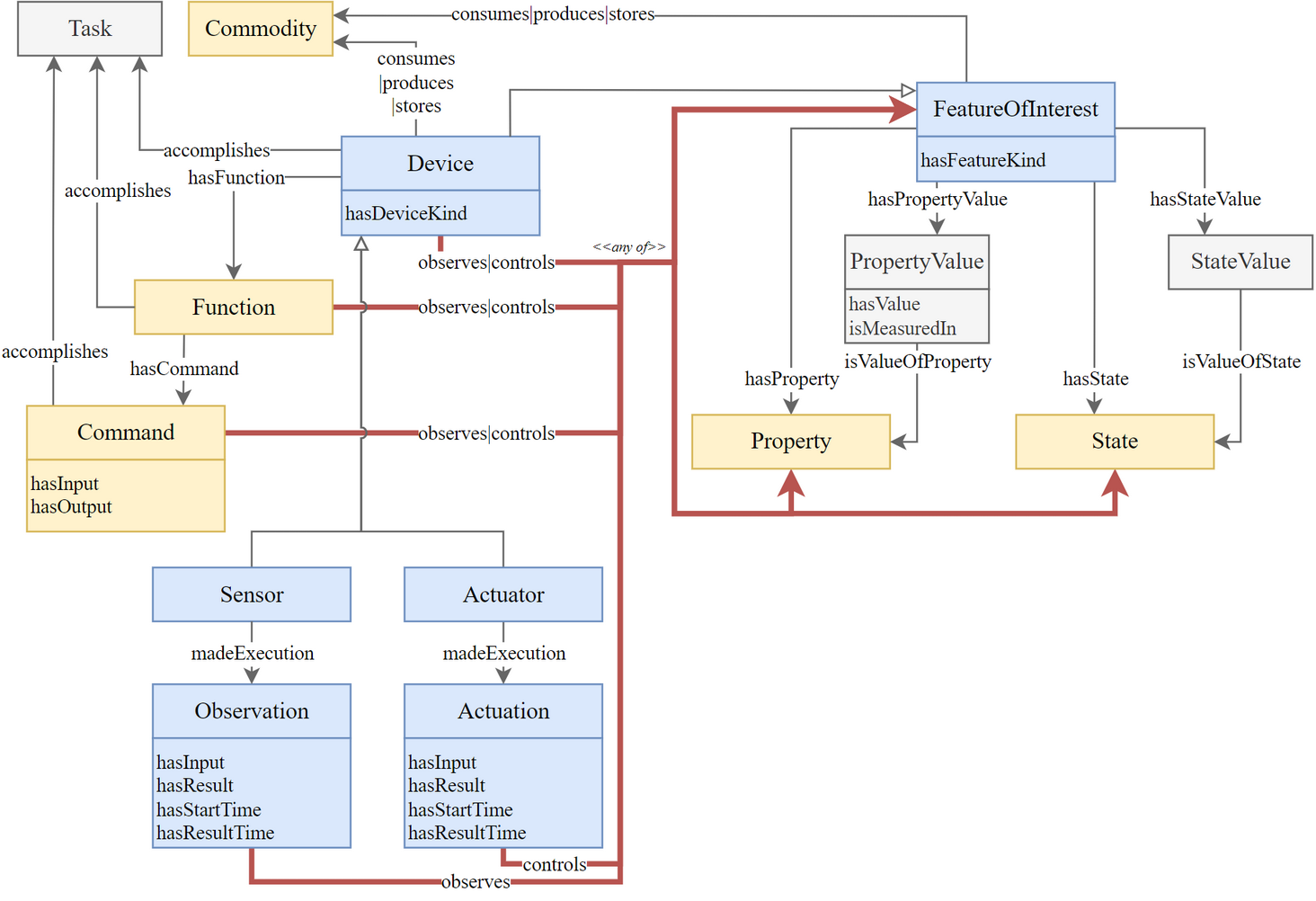}
    \caption[Overview of core SAREF ontology]{Overview of core SAREF ontology~\protect\footnotemark.}
    \label{fig:saref}
\end{figure*}
\afterpage{\footnotetext{Source: \url{https://saref.etsi.org/core/v4.1.1/}.}}

The SAREF Core ontology centres on devices as physical objects that perform tasks in buildings. For example, a washing machine can be represented as a device, while washing can be represented as a task. As shown in Figure~\ref{fig:saref}, SAREF models devices and their relation to their environment through classes, subclasses, object properties, and data properties. 
 Its main classes include
   {\small
   \mbox{\mul{sar}{saref:Device}},\,
   \mbox{\mul{sar}{saref:FeatureOfInterest}},\,
   \mbox{\mul{sar}{saref:Measurement}},\,
   \mbox{\mul{sar}{saref:Service}},\,
   \mbox{\mul{sar}{saref:Function}},\,
   \mbox{\mul{sar}{saref:Command}},\,
   and \mbox{\mul{sar}{saref:State}}.
 The SAREF documentation provides further details on these modelling elements~\cite{sarefcore}.

SAREF4ENER extends SAREF Core to represent device power profiles and energy-related scheduling concepts. Its main classes include \mul{s4b}{s4ener:PowerProfile}, \mul{s4b}{s4ener:PowerSequence}, \mul{s4b}{s4ener:AlternativesGroup}, and \mul{s4b}{s4ener:Slot}.  The SAREF4ENER documentation describes these scheduling and power-profile concepts in further detail~\cite{saref4ener}.

In SAREF4BLDG, SAREF Core is expanded with building devices, building objects, and building spaces. The extension draws on BOT and IFC concepts to represent building topology. Its main classes include \mul{s4b}{s4bldg:Building}, \mul{s4b}{s4bldg:BuildingSpace}, \mul{s4b}{s4bldg:PhysicalObject}, \mul{s4b}{s4bldg:BuildingObject}, and \mul{s4b}{s4bldg:BuildingDevice}.


SAREF4SYST is a cross-domain model for representing systems, system connections, and connection points. It comprises three main classes: \mul{s4b}{s4syst:System}, \mul{s4b}{s4syst:ConnectionPoint}, and \mul{s4b}{Connection}.  SAREF4GRID extends SAREF with concepts for IoT-oriented smart-grid data exchange~\cite{saref4grid}.

\subsubsection{\colorbox{ssn}{SSN}---Semantic sensor network and \colorbox{sos}{SOSA}---Sensor, observation, sample, and actuator}
\label{sec:ssn/sosa}
SSN/SOSA\phantomsection\label{ont:ssnsosa} combines two W3C ontologies: the Semantic Sensor Network ontology (SSN) and the Sensor, Observation, Sample, and Actuator ontology (SOSA). The W3C SSN Incubator Group developed SSN to describe sensor-based systems, including sensors, capabilities, measurement processes, observations, and deployments~\cite{compton2012ssn}. SOSA later emerged from a reconsideration of SSN in response to changes in scope, audience, technology, and lessons learned~\cite{janowicz2019sosa}. Together, SSN/SOSA provide a modular semantic framework for sensors, actuators, observations, procedures, samples, observed or actuated properties, and the systems to which they belong~\cite{haller2019modular}. It is often used to complement other building ontologies, as shown in Figure~\ref{fig:ssn/sosa}.

The main classes of SSN/SOSA represent concepts such as \mul{sos}{sosa:Actuator}, \mul{sos}{FeatureOfInterest}, \mul{sos}{sosa:Platform}, \mul{sos}{sosa:Observation}, \mul{ssn}{ssn:Deployment}, \mul{sos}{sosa:Procedure}, \mul{ssn}{ssn:System}, and \mul{sos}{sosa:Sensor}.

\begin{figure*}[ht!]
    \centering
    \includegraphics[width=0.95\textwidth]{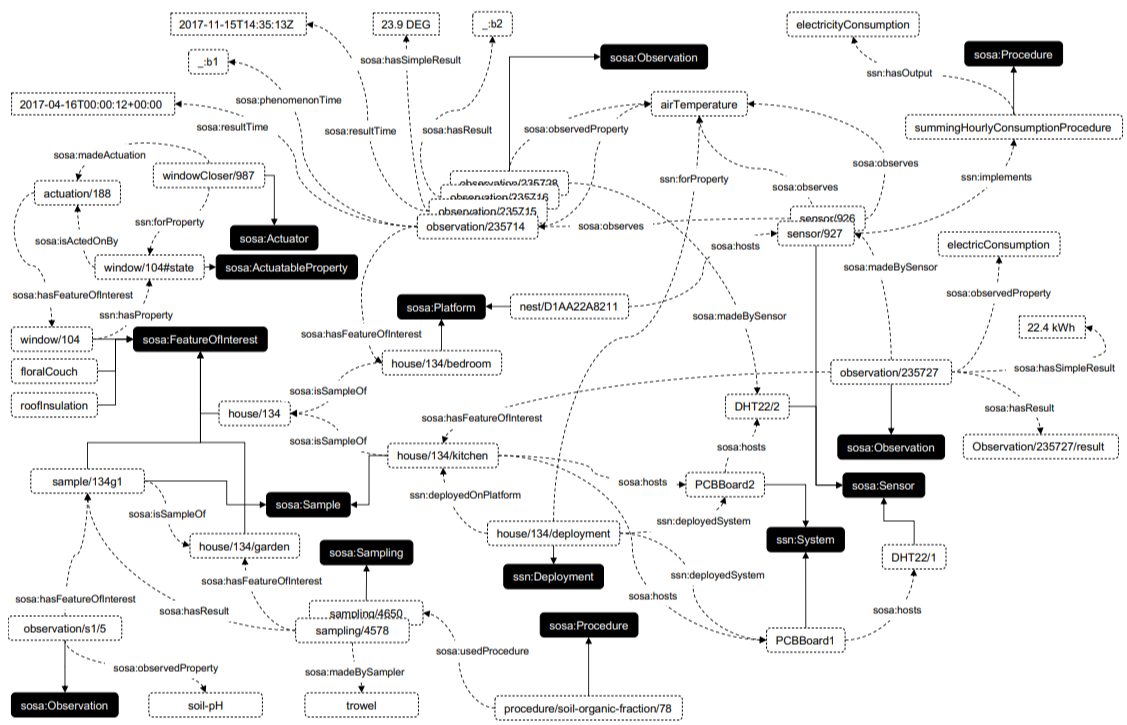}
    \caption{Integrating Building Data with Observation and Sensor Data Utilising SOSA/SSN (image from~\cite{haller2019modular}).}
    \label{fig:ssn/sosa}
\end{figure*}

\newpage
\subsubsection{\colorbox{bri}{Brick}}
\label{sec:brick}
Brick\phantomsection\label{ont:brick}~\cite{balaji2016brick} is an open-source semantic data modelling framework for smart buildings and building automation systems. It provides an extensible vocabulary for machine-readable descriptions of building assets, subsystems, points, locations, and relationships. Brick organises metadata around five core concepts: entity, tag, class, relationship, and graph, with entities represented through five high-level classes: \mul{bri}{brick:Collection}, \mul{bri}{brick:Equipment}, \mul{bri}{brick:Location}, \mul{bri}{brick:Measurable}, and \mul{bri}{brick:Point}.

In Brick, entities may be physical, virtual, or logical, covering equipment, lighting systems, meters, thermostats, rooms, floors, sensing points, energy-use points, actuation points, and operational groupings such as HVAC zones and lighting zones. Brick is structured as a directed, labelled graph; Figure~\ref{fig:brick} illustrates connections among an AHU, variable air volume boxes, an HVAC zone, rooms, and associated sensing and control points~\cite{luo2022extending}. Its design principles of completeness, expressiveness, usability, consistency, and extensibility support practical representation, consistent modelling, and extension when new building concepts are needed~\cite{balaji2016brick, balaji2018brick}.

\begin{figure*}[ht!]
    \centering
    \includegraphics[width=0.98\textwidth,height=0.42\textheight,keepaspectratio]{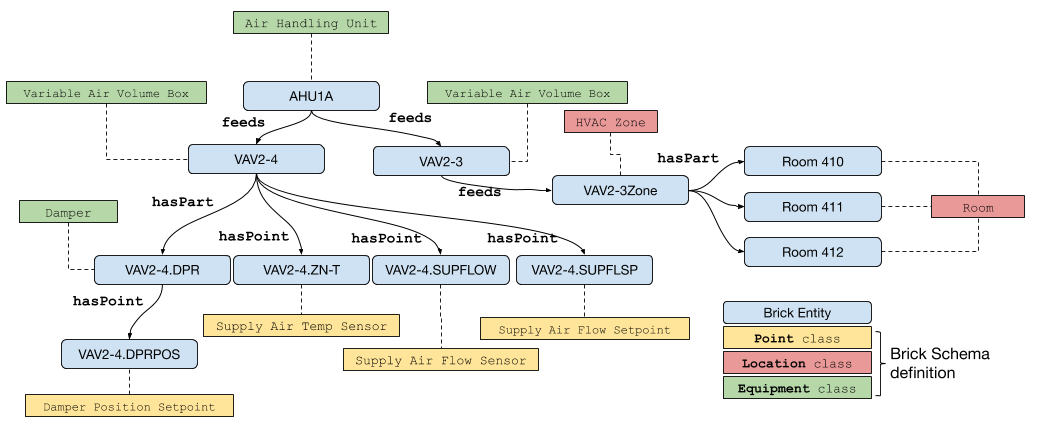}
    \caption{Brick model representing an AHU, two VAVs, and a handful of points and rooms (image from~\cite{brickOnline}).}
    \label{fig:brick}
\end{figure*}%

\subsubsection{\colorbox{oth}{PH}---Project Haystack}
\label{sec:project haystack}
PH\phantomsection\label{ont:ph} is a semantic data model for representing equipment and relationships in automation, control, energy, HVAC, and other environmental systems~\cite{projecthaystack}. It links physical building objects to entities, where an entity may represent a site, a piece of equipment such as an HVAC system or electric meter, or a point such as a digital or analogue sensor.

PH is organised around tags, each of which states a fact or attribute about an entity. For example, a \texttt{site} tag identifies an entity as a building or site, while \texttt{equip} and \texttt{point} tags identify equipment and data points. PH includes more than 200 tags for standardising descriptions of equipment, systems, and related data. Marker tags identify entity types without associated values, such as \texttt{ahu} for an air-handling unit, whereas string tags carry human-readable values.

\subsection{\colorbox{oth}{Other} domain-specific and supporting ontologies}

This section reviews additional domain-specific and supporting ontologies that complement the core models previously discussed. Rather than presenting these models only as isolated entries, the discussion groups them according to their main contribution to BEM: smart home and device modelling, building automation and control, energy flexibility and demand response, energy-system interoperability, and performance-oriented building analysis. Table~\ref{tab:other-ontology} provides the detailed catalogue of references, formats, class counts, scopes, and main focus areas, while the following descriptions highlight the main patterns across the reviewed models. These ontologies may be integrated with other models to represent building concepts not covered by a single ontology~\cite{ardjani2015ontology}. Ontology reuse refers to using an existing ontology as the basis for developing a new one~\cite{caldarola2016approach}. Although reuse and integration are related, distinguishing between them helps clarify how missing concepts are addressed in the use case studies discussed in Section~\ref{sec:usecases}.

\subsubsection{Smart home, device, occupancy, and ambient-intelligence ontologies}

In this group 
ThinkHome\phantomsection\label{ont:thinkhome}~\cite{reinisch2010thinkhome}, DogOnt\phantomsection\label{ont:dogont}~\cite{bonino2008dogont}, PowerOnt\phantomsection\label{ont:poweront}~\cite{bonino2015poweront}, and the Home Appliance ontology\phantomsection\label{ont:homeapplianceontology}~\cite{shah2011ontology} represent household energy use, devices, appliance-level consumption, comfort, user behaviour, and smart-home automation concepts.

The Home Appliance ontology is described as compatible with the Suggested Upper Merged Ontology (SUMO)~\cite{niles2001origins}. BonSAI\phantomsection\label{ont:bonsai}~\cite{stavropoulos2012bonsai} extends this direction towards service-oriented ambient intelligence by modelling operations, inputs, outputs, processing logic, parameters, and monitored or controlled environmental conditions using OWL-S~\cite{owl-s} and context-driven service adaptation concepts~\cite{preuveneers2004towards}.

Models concerned with occupants and comfort add a complementary perspective. ONCOM\phantomsection\label{ont:oncom}~\cite{orozco2019ontology} supports occupant emotional-state analysis for indoor thermal comfort, the Ambient Assisted Living ontology\phantomsection\label{ont:aalontology}~\cite{marcello2024digital} represents buildings, occupants, and connected household objects within an ontology-driven Digital Twin framework, and the Ontology Profile (OP\phantomsection\label{ont:op})~\cite{opontology} describes occupancy profiles and their influence on building energy use. Another occupant-behaviour ontology~\cite{hong2015ontology} standardises energy-related occupant behaviour in buildings, but is not included in Table~\ref{tab:other-ontology} because it relies only on XML and occupant behaviour XML format (obXML)~\cite{hong2015occupant}. Together, these models show that residential BEM applications often require semantic concepts for users, appliances, comfort, behaviour, and adaptive services.

\subsubsection{Building automation, control, and technical-system ontologies}
RealEstateCore\phantomsection\label{ont:realestatecore}~\cite{realEstateCore}, BACS\phantomsection\label{ont:bacs}~\cite{terkaj2017reusing}, BASont\phantomsection\label{ont:basont}~\cite{ploennigs2012basont}, and the Semantic BMS ontology (SBMS\phantomsection\label{ont:sbms})~\cite{kuvcera2018semantic} provide semantic support for smart buildings, building automation systems, device descriptions, data points, BIM-related elements, and operational analysis.

The TUBES System Ontology (TSO\phantomsection\label{ont:tso})~\cite{pauen2024integrated} adds topological, hierarchical, and functional modelling of building systems, especially system states in interconnected technical systems.

\changeurlcolor{blue}

\changeurlcolor{black}

\noindent Flow-oriented models such as FSO\phantomsection\label{ont:fso}~\cite{kukkonen2022ontology} and FPO\phantomsection\label{ont:fpo}~\cite{kucukavci2022taking} further represent energy and mass-flow relationships, component capacity, and technical system
properties associated with HVAC operation.

Models concerned with control and equipment extend this group towards operational deployment. The Building Control Ontology (CTRLont\phantomsection\label{ont:ctrlont})~\cite{schneider2017ontology} focuses on building automation control logic, ASHRAE 223P\phantomsection\label{ont:ashrae}~\cite{ashrae223p} standardises semantic formalism in building knowledge contextualisation, and Google Digital Buildings\phantomsection\label{ont:google digital buildings}~\cite{google, berkoben2020digital} offers structured representations of buildings and equipment. The Virtual Building Model metadata schema (VB schema~\phantomsection\label{ont:vbschema})~\cite{lee2025metadata} captures mathematical logic for the physical behaviour of buildings. It has been integrated with Brick and an AI-agent layer for the correction and calibration of building systems~\cite{yoon2025ontology}. Other smart-building models include SBonto\phantomsection\label{ont:sbonto}~\cite{vzavcek2017sbonto}, covering devices, states, architecture, environment, furniture, and networks, and Onto-SB\phantomsection\label{ont:ontosb}~\cite{degha2019intelligent}, supporting smart-building representation and reasoning for energy reduction. This group shows that operational BEM requires explicit descriptions of system composition, data points, control functions, and equipment behaviour.

\subsubsection{Energy flexibility, demand response, and grid-interactive ontologies}

A third group is concerned with energy flexibility, demand response, and interaction between buildings and energy systems. The Energy Flexibility Ontology (EFOnt\phantomsection\label{ont:efont})~\cite{li2022semantic} represents building energy flexibility resources and supports knowledge co-development for flexibility applications, while Mirabel\phantomsection\label{ont:mirabel}~\cite{verhoosel2012ontology} describes how devices express flexibility through user preferences and device energy profiles. SARGON\phantomsection\label{ont:sargon}~\cite{haghgoo2020sargon} extends SAREF4ENER with electrical-grid distribution and building energy automation concepts, and SARGON2\phantomsection\label{ont:sargon2}~\cite{tun2025sargon2} adds renewable energy resources, weather information, and power-system components.

Other models in this group focus more directly on demand response and grid-interactive operation. RESPOND\phantomsection\label{ont:respond}~\cite{respond} supports real-time optimal energy dispatch at dwelling, building, and district levels. OpenADR\phantomsection\label{ont:openadr}~\cite{fernandez2020openadr} standardises semantic representation in the demand response domain, DELTA\phantomsection\label{ont:delta}~\cite{fernandez2021supporting} models interactions between buildings and energy providers, and SG-BEMS\phantomsection\label{ont:sg-bems}~\cite{schachinger2016ontology} provides an abstraction layer for smart-grid and building interaction in BEMS operations. DEDALUS\phantomsection\label{ont:dedalus}~\cite{dedalus} supports demand response services in building-system interactions and extends the Demand Response Control Ontology (dCO\phantomsection\label{ont:dco})~\cite{dco}, which provides semantic foundations for devices, measurements, controls, and service interactions. These models highlight the growing need to represent flexibility resources, grid signals, and coordination mechanisms in BEM.

\subsubsection{Energy-system, smart-grid, and cross-domain interoperability ontologies}

A fourth group supports broader energy-system, smart-grid, and cross-domain interoperability. SEAS\phantomsection\label{ont:seas}~\cite{lefranccois2017seas} provides a common vocabulary for energy production, consumption, conservation, and efficiency, and can be aligned with domain ontologies such as SSN/SOSA. The Open Energy Ontology (OEO\phantomsection\label{ont:oeo})~\cite{booshehri2021introducing} supports energy-system modelling and analysis. Smart-grid models include SG\textsuperscript{1}\phantomsection\label{ont:sg1}~\cite{gillani2014generic}, describing a prosumer-oriented smart grid and supporting communication with buildings, and SG\textsuperscript{2}\phantomsection\label{ont:sg2}~\cite{tefek2023smart}, representing smart-grid entities and relationships for cyber-attack identification.
The Ontology-Based Information Model for Smart Grids (SSG\phantomsection\label{ont:ssg})~\cite{salameh2019ssg} is based on IFC and supports semantic interoperability among smart-grid components.

Several models in this group focus on integration across energy, weather, grid, and contextual data sources. SSGIM\phantomsection\label{ont:ssgim}~\cite{zhou2012semantic} integrates information for demand response applications using the International Electrotechnical Commission Common Information Model~\cite{ielectro}, DBpedia~\cite{dbpedia}, and a weather data model~\cite{weathermodel}. The Smart Electric Power Alliance ontology\phantomsection\label{ont:sepa's sg}~\cite{sepa} supports energy-consumption registration for smart-grid applications. OEMA\phantomsection\label{ont:oema}~\cite{cuenca2017unified} represents energy performance and contextual data across energy domains by drawing on ThinkHome, SAREF4ENER, SG\textsuperscript{1}~\cite{gillani2014generic}, and a deprecated energy-use ontology, while DABGEO\phantomsection\label{ont:dabgeo}~\cite{cuenca2020dabgeo} extends OEMA for energy management applications. The Common Information Model (CIM\phantomsection\label{ont:cim})~\cite{cim} represents smart-grid information within the Cerise-SG project. The Ontological Solution for Energy Intelligent Management~\cite{saba2019development} and NewOSEIM\phantomsection\label{ont:newoseim}~\cite{saba2021ontology} represent home-environment objects that affect electrical energy consumption and support reasoning for intelligent energy management. This group reflects the importance of semantic alignment across building, grid, weather, and energy-system data.

\subsubsection{Performance assessment, KPI, simulation, and building energy analysis ontologies}

A final group focuses on performance assessment, KPI representation, simulation interoperability, and building energy analysis. EEOnt\phantomsection\label{ont:eeont}~\cite{diaz2013eeont} represents building and energy-efficiency information for comparing building energy performance, while EEPSA\phantomsection\label{ont:eepsa}~\cite{esnaola2021eepsa} supports knowledge discovery in databases (KDD) processes and the identification of relationships among energy-efficiency data. The ontology-based framework introduced by \textit{Lork et al.}\phantomsection\label{ont:bem}~\cite{lork2019ontology} identifies causes of building energy efficiency and inefficiency. Models concerned with KPI representation include the Energy Management Key Performance Indicator ontology (EM-KPI\phantomsection\label{ont:em-kpi})~\cite{li2019enhancing}, which supports KPI exchange for districts and buildings, and the KPI ontology\phantomsection\label{ont:kpi}~\cite{kpiOntology}, which represents metrics associated with building renovation and energy-efficient building requirements.

Models concerned with performance and simulation add further analytical support. The Performance Assessment ontology (PF\phantomsection\label{ont:pf})~\cite{corry2015performance} supports granular building performance assessment and optimisation by drawing on SSN, SimModel~\cite{o2012simmodel}, a building simulation data model, and ifcOWL~\cite{pauwels2014representing}. While ifcOWL is included in Table~\ref{tab:other-ontology} as a foundational BIM linked-data model, SimModel is not included because it is primarily a simulation data model rather than an ontology commonly used in building operation studies. BOP\phantomsection\label{ont:bop}~\cite{donkersbuilding} integrates static and dynamic data points for thermal comfort analysis, SEMANCO\phantomsection\label{ont:semanco}~\cite{madrazo2012semanco} supports access to urban energy-related data, and EPG-RISK\phantomsection\label{ont:ero}~\cite{yilmaz2026developing} structures knowledge about energy performance gap risks. WoSO\phantomsection\label{ont:woso}~\cite{hounas2024web} extends SAREF to represent interactions between building performance simulations and IoT systems, SESAME\phantomsection\label{ont:sesame}~\cite{fensel2013sesame} integrates smart metering, building information, and rule-based reasoning for energy optimisation, BIGG\phantomsection\label{ont:bigg}~\cite{broto2025interoperable} integrates building physics, IoT, and time-series data while reusing SAREF, SSN, GEO~\cite{geo}, GeoNames~\cite{geonames}, VCARD~\cite{vcard}, and Time~\cite{timeOntology}, and ICBMS\phantomsection\label{ont:icbms}~\cite{ji2017intelligent} extends SAREF with state, sensor-measurement, and unit concepts for thermal comfort analysis. This group shows that analytical BEM applications increasingly require semantic support for KPIs, performance assessment, simulation links, time-series data, and operational interpretation.

This overview indicates that semantic modelling for BEM has expanded beyond the representation of physical building assets. Many models address advanced operational, analytical, and grid-interactive BEM functions. However, their diversity also illustrates the fragmentation of the ontology landscape. Several models overlap in scope, differ in openness and maintenance status, or are designed for narrow application contexts.
This reinforces the need for careful ontology selection, reuse, integration, and extension when applying semantic models to operational BEM use cases.

In relation to \emph{RQ2}, this review shows that semantic modelling for building operation is supported by a diverse ontology landscape rather than by a single dominant model. 
Core models such as BOT, SAREF, SSN/SOSA, Brick, and Project Haystack provide recurring foundations for representing topology, devices, observations, equipment, points, and operational metadata.

However, the additional ontologies reviewed in this section show that operational BEM also requires concepts related to automation, occupancy, energy flexibility, smart-grid interaction, performance assessment, and computational processes.
This motivates the use case analysis in the next section, which examines how these models are actually applied and combined in concrete BEM tasks.

\changeurlcolor{blue}
\section{Identifying ontology use case scenarios}
\label{sec:usecases}

 This section examines how semantic models are applied in building operational use cases, addressing \emph{RQ3} and preparing the discussion of \emph{RQ4}. The studies are organised according to single-ontology, two-ontology, and multiple-ontology strategies to show how models are selected, combined, or extended to satisfy the semantic requirements of specific BEM tasks. Beyond describing individual use cases, the section synthesises recurring patterns in ontology selection, semantic coverage, evidence traceability, and extension strategies.

\subsection{Ontology evidence completeness assessment}
\label{sec:oec_criteria}

The Ontology Evidence Completeness (OEC) is introduced in this survey as a classification framework to assess the traceability of semantic evidence in ontology-based use cases.
It makes the transparency and traceability of concept-to-class mappings explicit in ontology-based BEM studies. The assessment is based only on documented evidence available in the publications or associated materials, such as public repositories, ontology files, RDF or OWL serialisations, appendices, semantic diagrams, or explicit class-level mappings. OEC is categorised as \emph{Complete}, \emph{Partial}, or \emph{None}, indicating, respectively, whether class-level evidence is sufficient, incomplete, or unavailable for tracing task-relevant concepts to ontology classes.

To the best of the authors' knowledge, there is no known metric providing the same outcome as OEC.
Established ontology evaluation metrics are designed primarily to assess ontology artefacts, for example by measuring structural richness, logical consistency, competency-question coverage, domain coverage, or application performance~\cite{gomez2004ontology,hlomani2014approaches, duque2011oquare}. 
These metrics are valuable when the ontology, its requirements, mappings, and validation artefacts are available. 
However, the purpose of OEC is not to judge the intrinsic quality or completeness of an ontology, nor to measure how well an ontology represents domain concepts. 
Instead, OEC assesses how transparently an applied study reports the semantic evidence needed to trace task-relevant operational concepts to the ontology classes used to represent them.
This distinction is necessary because many reviewed studies mention ontology use without providing the class mappings, RDF/OWL files, repositories, appendices, or semantic diagrams needed to inspect concept-to-class representations. The absence of such evidence may also affect the reliability of the analyses conducted.

Table~\ref{tab:oec_criteria} summarises the OEC categories, the evidence inspected, and illustrative examples used to guide the classification in this study.
\begin{table*}[!t]
\caption{Ontology evidence completeness assessment criteria used in the use case analysis.}
\label{tab:oec_criteria}
\centering
\footnotesize
\setlength{\tabcolsep}{3pt}
\renewcommand{\arraystretch}{1.08}
\begin{tabularx}{\textwidth}{@{}>{\RaggedRight\arraybackslash}p{1.45cm} *{4}{>{\RaggedRight\arraybackslash}X}@{}}
\toprule
\makecell[l]{\textbf{OEC}\\\textbf{category}} & \textbf{Assessment criterion} & \textbf{Evidence inspected} & \textbf{Example study} & \makecell[l]{\textbf{Reason for}\\\textbf{classification}} \\
\midrule
\textbf{Complete} & The main task-relevant concepts can be traced to ontology classes. & Explicit class mappings, ontology files, RDF/OWL serialisations, appendices, or semantic-layer diagrams. & Smart home data integration and temperature-control use case~\cite{van2020validating}. & The study identifies the SAREF classes used for the required smart-home concepts. It also provides a table of the modelled concepts and a GitHub repository containing the implementation. \\
\addlinespace[1pt]
\textbf{Partial} & Some task-relevant concepts can be traced to ontology classes, but the mapping is incomplete. & Partial semantic diagrams, selected class examples, incomplete repository material, or query patterns showing only part of the semantic model. & Semantic infrastructure for building system operations~\cite{delgoshaei2022semantic}. & The study demonstrates ontology use through rules, queries, and application scenarios. Some concepts can be traced to classes from BOT, SAREF4BLDG, SOSA, Brick, QUDT, and Time, but the study does not provide a complete concept-to-class mapping for the full framework or an open repository for further inspection. \\
\addlinespace[1pt]
\textbf{None} & The ontology is mentioned, but class-level evidence is insufficient to inspect concept-to-class mappings. & General statements that an ontology was used, without ontology files, serialisations, diagrams, appendices, or explicit class mappings. & AHU fault diagnosis using PH and Brick~\cite{nehasil2021versatile}. & The study reports using PH and Brick, but does not provide sufficient class-level mappings for the quantitative concept-to-class analysis. \\
\bottomrule
\end{tabularx}
\end{table*}
Using these criteria, the following analysis reports qualitative use-case evidence and quantitative concept-to-class coverage.

Table~\ref{tab:usecase} summarises the reviewed use case scenarios, the ontologies employed, and the ontology classes explicitly reported or traceable from each study. OEC was manually assessed by inspecting the publication and, where available, associated repositories, ontology files, appendices, diagrams, or triple serialisations.
The analysis focuses on concept-to-class coverage because ontology classes are more consistently reported across the reviewed studies and provide a more comparable basis for evaluating semantic coverage.
This summary does not assume a one-to-one correspondence between concepts and classes, since operational concepts may require multiple classes and individual classes may support multiple concepts. General-purpose upper ontologies for time, units, or spatial location are treated as supporting resources rather than complementary BEM ontologies.

Table~\ref{tab:quantitative} provides a quantitative synthesis of the reviewed use cases, reporting how extensively each selected ontology was applied and the extent to which additional modelling was required.
The first column identifies the use case and shows the corresponding OEC class.
The \textit{Number of Concepts} column denotes the total number of classes available in the selected ontology and, where applicable, the parent-class denominator used to compute the OIR. The notation \(P\) denotes the number of parent classes considered for the denominator. \textit{Used Concepts} reports how many task-relevant concepts were directly supported by existing ontology classes. The \textit{Number of Concepts Created} column captures additional classes introduced to represent required concepts not covered by the selected ontology.
The quantitative analysis uses two complementary metrics: Ontology Instantiation Rates (OIR) and Necessity to Extend (NTE). The OIR is computed with respect to the ontology's parent class structure and is defined as the proportion of parent classes used in the use case, where a parent class is counted if it supports at least one required concept.
The total number of parent classes was determined from the ontology structure using ontology inspection tools, such as Ontospy, as described in Section~\ref{sec:semantic models}. The metric therefore reflects the breadth of ontology use at the parent-class level, rather than the proportion of all ontology classes instantiated or the overall conceptual coverage of the ontology. The NTE quantifies the proportion of required concepts that required newly created classes. The method of concept coverage identifies how missing concepts were addressed, including ontology class specialisation, ontology external inheritance, ontology reuse, ontology integration, and application ontology extension.
The method of concept coverage distinguishes ontology class specialisation, ontology external inheritance, ontology reuse, ontology integration, and application ontology extension. These categories capture whether missing concepts are addressed by specialising existing classes, importing external classes, modifying an ontology, combining multiple ontologies, or adding application-specific concepts in a parallel namespace.

For multiple-ontology integration cases, the reported values are ontology specific. \textit{Used Concepts} indicates concepts directly represented by the ontology listed in the corresponding table entry, whereas \textit{Number of Concepts Created} indicates concepts not directly represented by that ontology and addressed through the other ontologies or extensions used in the integrated model. These values therefore describe ontology-specific coverage gaps within an integrated modelling strategy and should not be read as separate, independent concept sets for each ontology. 
These metrics are defined as follows:
\begin{align}
\text{OIR (\%)} &= \frac{P_{\text{used}}}{P_{\text{total}}} \times 100 \\[3pt]
\text{NTE (\%)} &= \frac{C_{\text{created}}}{C_{\text{used}} + C_{\text{created}}} \times 100
\end{align}
\noindent where \(P_{\text{used}}\) is the number of parent classes instantiated/used to represent the task-relevant concepts, \(P_{\text{total}}\) is the total number of parent classes in the selected ontology,
\(C_{\text{used}}\) is the number of task-relevant concepts covered by existing ontology classes \textit{(Used Concepts)},
\(C_{\text{created}}\) is the number of task-relevant concepts not covered by the ontology and therefore represented through newly created classes \textit{(Number of Concepts Created)}, and
\(C_{\text{total}}\) is the \(C_{\text{used}} + C_{\text{created}}\) = total number of task-relevant concepts identified in the study.

The following subsections apply this framework to the reviewed use cases, linking the extracted evidence to the coverage and modelling strategies reported in Table~\ref{tab:quantitative}.

}

\subsection{Single-ontology use case scenarios}
Single-ontology use cases refer to data-driven applications that rely primarily on one ontology, or one ontology family, to describe building operational data and its contextual meaning. This section examines how such models are used in specific BEM tasks and how much semantic evidence is available for their reported concept coverage.

This study proposed an ontology-based energy modelling framework for scalable and adaptable building digital twins~\cite{bjornskov2023ontology}. Within building operations, the use case addressed the dynamic simulation of single-zone building systems and components. OEC is Complete because the study explicitly identified the ontology classes used to represent the concepts required for the building energy simulation task.

The study used SAREF, SAREF4BLDG, and SAREF4SYST to represent spatial, HVAC, sensing, control, and indoor environmental concepts, while gaps remained for outdoor conditions, schedules, node-level flow representation, and solar irradiation. The OIRs were 25\%, 50\%, and 100\%, respectively, and the remaining 26.67\% coverage gap was addressed through ontology class specialisation. Subsequent studies extended the approach to larger building contexts and building performance analysis~\cite{bjornskov2025large, bjornskov2025automated}.

The study validated SAREF in a smart home environment by mapping device data into RDF and using the ontology to support interoperability, room-condition monitoring, and automated temperature control for thermal comfort~\cite{van2020validating}.
For this operational task, SAREF represented the main device, measurement, property, comfort, and control concepts, while the OM1.8 Unit of Measure ontology~\cite{rijgersberg2013ontology} was used for measurement-unit concepts not adequately covered by SAREF. OEC is Complete because the study explicitly identified the SAREF classes used for the task-relevant concepts.
The OIR was 45\%, and the NTE was 16.67\%. This gap was addressed through ontology external inheritance by incorporating the relevant unit-of-measurement classes from the upper ontology into the SAREF namespace.

Using the Open Power System Data (OPSD) household dataset~\cite{opsd}, SAREF was applied to harmonise smart home energy data and support occupant detection from energy consumption patterns~\cite{weerdt2021making}.
This use case focused on electricity-related observations, including grid exchange, photovoltaic generation, electric vehicle charging, and appliance-level consumption. OEC is Complete because explicit SAREF class mappings were provided for the required device, measurement, property, metering, and appliance concepts, while the kilowatt-hour unit was represented through OM1.8.
The OIR was 30\%, and the NTE was 11.1\%. As in the preceding SAREF use case, the gap was addressed through ontology external inheritance, although here the extension need was limited to the unit-of-measurement concept.

OfficeGraph extends the SAREF-oriented smart home examples towards a larger office-building setting, where heterogeneous IoT measurements were organised as a real-world knowledge graph for building management, analytics, and machine learning~\cite{van2024officegraph}.
The use case reused SAREF, SAREF4BLDG, SAREF4ENER, and OM 1.8 to represent core sensing, measurement, unit, and building-space concepts, while additional graph elements were introduced for application-specific analytics. OEC is Partial because the class-level evidence covers the core semantic model but remains incomplete for parts of the graph-based workflow.
Based on the documented class-level evidence, the OIR is 15\% for SAREF Core, 12.5\% for SAREF4BLDG, and 2.12\% for SAREF4ENER, while the NTE is 37.5\%. Missing concepts were addressed mainly through application ontology extension by introducing concepts in the parallel \textit{ic}: namespace associated with the Interconnect project~\footnote{\url{https://interconnectproject.eu/}}, alongside the reused SAREF-based ontologies.

A further SAREF-centred application converted EnergyPlus simulation outputs and IoT data into a knowledge graph to support advanced querying, reasoning, and energy optimisation in buildings~\cite{kirnapci2025ontology}.
The semantic layer used SAREF as the main ontology backbone and OM for unit representation, covering indoor conditions, occupancy, sensing, room-level measurement context, and measurement units. OEC is Partial because core measurement and occupancy concepts are mapped to classes, while energy-use, control, and optimisation concepts remain insufficiently documented at class level.
From the available class-level evidence, the OIR is 30\% for SAREF Core, and the NTE is 33.33\%. The missing concepts were handled through ontology external inheritance.

The next use case shifts from SAREF to Brick, using semantic modelling to support scalable deployment of AI-enabled, data-driven building management applications across different buildings~\cite{xu2024semantic}.
The demonstrated application focused on cooling-load prediction, where Brick-based descriptions supported building-independent data acquisition and deployment by representing equipment, location, sensing, and time-series related concepts.
OEC is Partial because the framework illustrates the use of Brick classes in the semantic model and query process, but does not provide a complete concept-to-class mapping for every concept required by the cooling-load prediction task. On the basis of Brick's five parent classes, the OIR was 60\%.
Although no major missing concept was explicitly defined for the demonstrated use case, the study notes that more advanced applications, including building energy flexibility analysis, would require semantic support beyond the current Brick-based framework. Ontology integration was therefore identified as the relevant strategy for broader extensibility.

Brick was also used in a portable EMIS workflow that combines semantic modelling and machine learning for photovoltaic-system anomaly detection across multiple campus sites~\cite{chiosa2024portable}. The task required descriptions of photovoltaic assets, electrical power measurements, weather variables, location, and time-series identifiers. OEC is Partial because core photovoltaic and sensing concepts are represented, while supporting workflow metadata, including geographical coordinates, campus-relative location, and data-retrieval identifiers, is not documented with the same class-level detail. The OIR is 60\%, reflecting the use of Brick Equipment, Point, and Location; since no explicit concept-level gaps were reported, the NTE is 0\%.

Another Brick-based use case addressed interoperability between proprietary building energy modelling data and linked building data applications~\cite{zheng2023towards}.
Here, Brick was used as the core ontology and extended to represent BEM concepts that were not directly covered, especially building-envelope elements, geothermal plant equipment, and occupant behaviour. The resulting framework brought together spatial, envelope, HVAC, control, sensing, actuation, and occupant-related concepts so that BEM data could be analysed together with graph and time-series data sources.
The OEC classification is Partial because many required concepts are explicitly mapped to ontology classes, covering spatial hierarchy, building systems, control and sensing points, envelope elements, HVAC and plant equipment, and occupant-behaviour concepts. However, the study does not provide explicit class-level mappings between simulated time series and ontology entities, even though the time-series data are stored in a temporal database.
The OIR is 60.0\% for Brick, based on the documented use of the Location, Equipment, and Point parent classes, while 42.10\% of the required concepts remained outside Brick's direct coverage. These gaps were addressed through ontology class specialisation.

For long-term thermal comfort evaluation, Brick was used to make the application portable and reproducible across heterogeneous buildings~\cite{sun2022enabling}.
The evaluation relied on the Mortar testbed~\cite{fierro2019mortar}, which contains data from more than 25 buildings and enabled the generalisability of the application to be tested beyond a single site.
The semantic requirements focused on spatial hierarchy, air-distribution equipment, and zone air-temperature sensing through concepts such as zones, rooms, floors, air handling units, variable air volume boxes, and zone air temperature sensors.
OEC is Complete because the ontology classes used for the task-relevant concepts are explicitly specified, and no missing concepts were identified. The OIR was 60\%; since no concept gap was reported, no extension strategy was required.

In a digital-twin-oriented BEMS application, Brick was used as the base semantic model for an ontology supporting real-time energy efficiency analysis~\cite{hwang2025dt}.
The semantic layer represented control points and data-collection processes, while three behavioural models supported energy imputation, occupancy detection, and thermal environment assessment using the Predicted Mean Vote (PMV) index.
The required concepts covered spatial context, building function, office space, HVAC and photovoltaic systems, energy use and generation, and indoor environmental variables such as temperature, humidity, and carbon dioxide concentration.
OEC is Partial because the study presents a Brick-based BEMS ontology for the target space, but does not provide complete concept-to-class mappings for all concepts required by the wider digital-twin BEMS workflow.
Based on Brick's parent-class structure, the OIR is 60\%, reflecting the use of Location, Equipment, and Point. Since no explicit concept-level gaps were reported, the NTE is 0\% and no extension strategy is identified.

A broader Brick use case considered how metadata from different stages of the building lifecycle can be normalised into a unified representation for data-driven analytics and operational applications~\cite{fierro2020shepherding}.
\noindent The work focused on cross-source semantic alignment for HVAC-related building systems, covering locations, spaces and thermal zones, equipment and systems, sensing and monitoring points, and working fluids such as air and water.
OEC is Partial because the repository 
shows implementation-level incompleteness, including unresolved mappings for some source elements and ambiguous mappings for certain domain components.
Even with these limitations, the implementation used four out of five core top-level class families, giving an OIR of 80\% and indicating substantial semantic support for the operational task. Within the scope of the study, no missing concepts directly relevant to the task were identified. Therefore, the NTE is 0\% and no semantic extension strategy was required.

Grid-responsive supervisory control provides a different single-ontology use case, in which open semantic requirements were used to configure portable demand-flexibility applications in buildings~\cite{paul2025open}.
The demonstrated application implemented two demand-flexibility functions for shifting HVAC load in response to grid conditions, including temperature setpoint adjustment, reducing coincident demand peaks, and coordination of zone-level HVAC units across simulated and real buildings.
ASHRAE Standard 223P was used with Shapes Constraint Language (SHACL)~\cite{shacl} to define machine-readable semantic requirements, validate building models, and support semi-automated application configuration.
OEC is Partial because some classes used to represent building systems are explicitly identified, but the complete semantic model is not available for examination. Consequently, the OIR cannot be computed, and the study is excluded from Table~\ref{tab:usecase} and Table~\ref{tab:quantitative}.

Overall, the single-ontology use cases show that individual models can support operational BEM tasks when the required concepts remain close to their original modelling scope. SAREF is mainly applied to device, measurement, property, and smart-home interoperability scenarios, whereas Brick is mainly applied to building equipment, points, locations, and operational metadata. The recurring finding is that single ontologies provide useful semantic foundations, while additional modelling strategies are often used to address concepts beyond their original scope, such as units, schedules, contextual data, analytics, or task-specific decision-support concepts. Other single-ontology studies report applications in cloud-based management services, building performance monitoring, anomaly detection, and optimal control of building systems~\cite{garcia2023etsi, cubides2025data, yin2025knowledge, thomsen2025ontology}. These studies are not included in Table~\ref{tab:usecase} or Table~\ref{tab:quantitative} because their OEC is classified as None, meaning that the available documentation does not provide sufficient class-level evidence for the quantitative concept-to-class analysis.

\subsection{Two-ontology use case scenarios}
This section identifies two ontology use cases, 
including those that combine existing ontologies to represent additional concepts within a new ontology.

The first two-ontology use case concerns a hybrid inference system for estimating indoor environmental conditions from real-time BAS data collected in the Researcher Hotel at Aalto University, Espoo, Finland~\cite{ji2017intelligent}.
The operational task focused on apartment-level thermal comfort assessment using temperature, humidity, occupancy status, and CO$_2$ concentration.

OEC is Complete because the study explicitly illustrates how the required sensing, state, and comfort-related concepts are represented in the semantic model.
SAREF and SAREF4BLDG provided partial support for the use case, with OIRs of 10\% and 12.5\%, respectively, but they did not adequately cover thermal comfort states, occupancy state levels, PMV and predicted percentage dissatisfied indices, or the unit of measurement required for the CO$_2$ sensor.

The resulting NTE was 80.95\%. To address these gaps, the ICBMS (IoT, Cloud, Big Data, Mobile, Security) ontology was introduced by reusing and extending SAREF to represent the missing concepts.

Unit-related concepts were further handled through ontology external inheritance by incorporating upper ontologies such as the Time ontology~\cite{timeOntology} and the Ontology of Units and Measure~\cite{rijgersberg2013ontology}.

Semantic BMS provides another example of creating a more domain-specific ontology by extending a general sensor ontology for building operation analysis~\cite{kuvcera2018semantic}.
The middleware framework enriched building automation data with structured semantics and linked them to BIM-related information, supporting facility benchmarking, building performance evaluation, decision support, and BMS data querying.

Its operational scope included room-environment assessment and energy-consumption analysis, with concepts covering spatial entities, devices, sensors and actuators, data points, observations, measured properties, sensing types, HVAC signals, energy consumption, room conditions, and BIM-derived building elements.
OEC is Complete because the thesis explicitly documents the ontology structure, class-level specialisations, and their role in the implemented querying framework.

The semantic model was grounded in SSN, which had an OIR of 19.05\%, and was then extended through the SBMS ontology, increasing the OIR to 55.55\%.
The NTE was 64.28\%, reflecting the inability of SSN alone to represent building-automation concepts such as data points, BMS addresses, structured property domains, and BIM-linked entities. These limitations were addressed through ontology reuse and ontology class specialisation in the development of SBMS.

A demand-flexibility control application illustrates a different form of two-ontology use, where Brick and SAREF were aligned to support automated control of flexible HVAC loads under both building and grid conditions~\cite{de2025semantics}.
Brick and SAREF were aligned and integrated to support demand-flexibility control concepts spanning building zones, sensing, setpoints, and grid interaction. OEC is Complete because the study specified the ontology classes used for the key concepts and identified unsupported concepts. 
The reported mappings show that Brick provided more direct class-level support for several operational concepts, whereas SAREF often required combinations of classes.

For the demonstrated task, the OIRs were 60.0\% for Brick, 15\% for SAREF Core, and 50.0\% for SAREF4BLDG. 
Missing concepts were identified in SAREF, particularly for modelling certain device properties, giving a NTE rate of 7.35\%.
QUDT (Quantities, Units, Dimensions and Types)~\footnotemark{} was additionally used through ontology external inheritance to represent property concepts, particularly temperature and occupancy, within the SAREF observation and control patterns.\footnotetext{\url{http://www.qudt.org/}} Grid signals remained unsupported and were not modelled in the demonstrated use case, although representation through OpenADR was suggested but not implemented.  
Overall, the gaps were addressed through ontology integration and ontology external inheritance.

A diagnostic use case based on IBM's Semantic Smart Building Diagnoser used semantic sensor network models to derive diagnosis rules for smart buildings~\cite{ploennigs2014adapting}.
The operational task was fault detection and diagnosis of atypical building behaviour, such as unusually high room temperature, by tracing possible causes through physical process links between sensors, properties, actuators, and building conditions.
The use case required semantic support for sensing, building conditions, actuation, anomaly states, and causal diagnostic relationships. SSN provided the backbone ontology, but additional physical-process and smart-building concepts were needed to represent cause-effect mechanisms for diagnosis.
OEC is Partial because the paper describes the ontology classes used in the core diagnosis model and illustrates them through conceptual models and a case study, but does not provide a fully exhaustive concept-to-class mapping for the complete building-scale diagnostic deployment.
SSN parent classes accounted for 23.81\% of the relevant concept representation, while the NTE was 90.48\%. The missing concepts were addressed by developing a new ontology on the basis of SSN through ontology reuse and ontology class specialisation, resulting in an OIR of 85.71\%.

\onecolumn
\begin{longtblr}[
  caption = {Usecase Scenarios on the Application of Ontologies.},
  label = {tab:usecase},
]{
  width   = \textwidth,
  colspec = {Q[0.72,l,m] Q[1.38,l,m] Q[2.20,l,m] Q[1.50,l,m]},
  colsep  = 1.5pt,
  cells   = {font = \scriptsize\selectfont},
  rowhead = 1,
  hlines,
}
\cline[2pt]{-}
\textbf{Usecase} & \textbf{Concepts} & \textbf{Classes} & \textbf{Description}\\
\SetCell[c=4]{c, bg=s4b, font=\footnotesize\bfseries} Single-ontology use case scenarios & & & \\
Data Integration for building energy analysis~\cite{bjornskov2023ontology} & Building Space, Building system and components, CO$_2$ sensor and controller & Uses SAREF, S4BLDG, S4SYST \cvtag{s4b}{BuildingSpace}
\cvtag{sar}{SpaceHeater}
\cvtag{s4b}{Valve}
\cvtag{sar}{Sensor}
\cvtag{s4b}{Controller}

\cvtag{s4b}{Damper}
\cvtag{s4s}{System}
\cvtag{s4s}{Connection}
\cvtag{s4s}{ConnectionPoint}

& SAREF ontology was incorporated to model a single zone system. \\

Data Integration in Smart Homes~\cite{van2020validating} & Current temperature, CO$_2$ sensor,
First time movement detected, Last time movement detected, Motion detected, Presence detected, Humidity, Smart switch actuator, Target temperature
& Uses SAREF
\cvtag{sar}{Measurement}
\cvtag{sar}{UnitofMeasure}
\cvtag{sar}{TemperatureUnit}
\cvtag{sar}{Property}
\cvtag{sar}{FeatureOfInterest}
\cvtag{sar}{Device}
\cvtag{sar}{Task}
\cvtag{sar}{Comfort}
\cvtag{sar}{WellBeing}
\cvtag{sar}{Function}
\cvtag{sar}{SensingFunction}
\cvtag{sar}{Command}
\cvtag{sar}{State}
\cvtag{sar}{Service}
& SAREF does not possess a class specifically designed to model the unit of measurement for CO$_2$, nor a property class for measuring CO$_2$levels. Nonetheless, it can be easily extended to include these features.\\

Data Integration with OPSD Household Data set~\cite{weerdt2021making} & Energy imported/exported from/to the public grid, Total photovoltaic energy generation, Electric vehicle Charging energy, Appliances and circulation pump energy consumption & Uses SAREF
\cvtag{sar}{Measurement}
\cvtag{sar}{UnitOfMeasure}
\cvtag{sar}{Property}
\cvtag{sar}{PowerUnit}
\cvtag{sar}{Power}
\cvtag{sar}{FeatureOfInterest}
\cvtag{sar}{Device}
\cvtag{sar}{Meter}
\cvtag{sar}{Appliance}
\cvtag{sar}{Task}
\cvtag{sar}{MeterReading}
\cvtag{sar}{Washing}
\cvtag{sar}{WellBeing}
\cvtag{sar}{Comfort}
\cvtag{sar}{Function}
\cvtag{sar}{MeteringFunction}
& The SAREF ontology was incorporated to map building data using the OPSD Household data set. \\

A Knowledge Graph of Office Building IoT Measurements \newline(OfficeGraph)\newline~\cite{van2024officegraph} & device, property, measurement, unit of measurement, battery level, CO2 level, device status, running time, thermostat heating setpoint, humidity, motion, occupancy, power, temperature, and building space & Uses SAREF and \textit{ic} namespace from Interconnect H2020 Project:
\cvtag{s4e}{Device}
\cvtag{sar}{UnitOfMeasure}
\cvtag{sar}{Property}
\cvtag{sar}{Measurement}
\cvtag{s4b}{BuildingSpace}
\cvtag{oth}{ThermostatHeatingSetpoint}
\cvtag{oth}{RunningTime}
\cvtag{oth}{Contact}
\cvtag{oth}{BatteryLevel}
\cvtag{oth}{CO2Level}
\cvtag{oth}{DeviceStatus}
& Integration of heterogeneous IoT device data and application of machine learning experiments to generate building insights.
\\

Ontology-Driven Smart Building Semantics \newline ~\cite{kirnapci2025ontology} & Temporal information, environmental conditions, indoor measurements, occupancy, energy use of window fan, HVAC energy consumption, HVAC setpoints, Air conditioning fan speed, and window fan speed & Uses SAREF:
\cvtag{sar}{Task}
\cvtag{sar}{Device}
\cvtag{sar}{Property}
\cvtag{sar}{Measurement}
\cvtag{sar}{UnitOfMeasure}
\cvtag{sar}{FeatureOfInterest}  & Integration of simulation and IoT data using SAREF to support building operational tasks.
\\

Semantic modeling for AI-driven building management applications \newline ~\cite{xu2024semantic} & Building system and equipment components, spatial context, sensors and point measurements, temporal information, thermal variables, and weather-related information & Uses Brick
\cvtag{bri}{Point}
\cvtag{bri}{Equipment}
\cvtag{bri}{Location} & The adoption of building semantics to enable standardised data queryability and the portability of BEM operational tasks, as demonstrated in cooling load prediction
\\

Energy management and information systems \newline ~\cite{chiosa2024portable} & Outdoor air temperature, Solar irradiance, Electrical power, and Location & Uses Brick
\cvtag{bri}{temperature Sensor}
\cvtag{bri}{Active Power Sensor}
\cvtag{bri}{Solar Radiance Sensor}
\cvtag{bri}{PV Generation System}
\cvtag{bri}{Weather Station}
\cvtag{bri}{Site} & Anomaly detection in Photovoltaic System

\\
Towards linked building data \newline ~\cite{zheng2023towards} & Space, Zone, Building, Floor, Envelope, System, Equipment, HVAC equipment, Lighting equipment, Control device, Control point & Use Brick
\cvtag{bri}{Point}
\cvtag{bri}{Equipment}
\cvtag{bri}{Location} & A data framework for enabling building systems interoperability through an extended Brick ontology

\\
Long-term Thermal Comfort Evaluation with Brick Schema\newline~\cite{sun2022enabling} & Air Handling Unit, Variable Air Volume box, Zone Air Temperature, Room and Floor & Uses Brick
\cvtag{bri}{Zone}
\cvtag{bri}{Room}
\cvtag{bri}{Floor}
\cvtag{bri}{AHU}
\cvtag{bri}{Variable Air Volume Box}
\cvtag{bri}{Zone Air Temperature Sensor}

&  The Brick ontology was employed to formalise specific building entities to analyse thermal comfort evaluation.\\

Ontology-based building energy management system ~\cite{hwang2025dt}& Location, Building usage, Office space, HVAC system, and Photovoltaic system & Uses Brick
\cvtag{bri}{Location}
\cvtag{bri}{Equipment}
\cvtag{bri}{Point}  &

The Brick ontology was used within the digital twin layer to support information fusion and data-driven processes for energy efficiency evaluation.

\\

Data integration for data-driven analytics~\cite{fierro2020shepherding} & Building system and equipment, Data points & Uses Brick
\cvtag{bri}{Air Handling Unit}
\cvtag{bri}{Power Meter}
\cvtag{bri}{Building Power Meter}
\cvtag{bri}{Rooftop Unit}

\cvtag{bri}{VAV}
\cvtag{bri}{Chiller}
\cvtag{bri}{Absorption Chiller}
\cvtag{bri}{Temperature Sensor}
\cvtag{bri}{Flow Sensor}
\cvtag{bri}{Return Air Flow Sensor}
& Brick served as a unified model for integrating metadata from different data sources.\\

\SetCell[c=4]{c, bg=s4b, font=\footnotesize\bfseries} Two-ontology use case scenarios & & & \\

Energy and Thermal Comfort Analysis~\cite{ji2017intelligent} & Thermal Comfort State, Number of occupants, Occupancy State, Building Space, Humidity  and CO$_2$Sensor, PMV and PPD values & Uses ICBMS

\cvtag{icb}{CO$_2$Sensor}
\cvtag{icb}{TemperatureSensor}
\cvtag{icb}{HumiditySensor}
\cvtag{icb}{State}
\cvtag{icb}{ThermalComfortState}
\cvtag{icb}{HotState}
\cvtag{icb}{WarmState}
\cvtag{icb}{SlightlyWarmState}
\cvtag{icb}{NeutralState}
\cvtag{icb}{SlightlyCoolState}
\cvtag{icb}{CoolState}
\cvtag{icb}{ColdState}
\cvtag{icb}{OccupancyState}
\cvtag{icb}{OccupiedState}
\cvtag{icb}{VacantState}
\cvtag{sar}{BuildingSpace}
\newline Extends SAREF
\cvtag{sar}{Sensor}
\cvtag{sar}{State}
& ICBMS ontology extends the SAREF ontology and introduces more specific concepts relating to sensor and state information \\

Data integration for building operational analysis~\cite{kuvcera2018semantic} & Data Points, Sensing, Source, Property, and Feature of Interest & Uses SBMS
\cvtag{sbm}{Address}
\cvtag{sbm}{DataPoint}
\cvtag{sbm}{Input}
\cvtag{sbm}{Property}
\cvtag{sbm}{FeatureOfInterest}
\cvtag{sbm}{SensingDevice}
\cvtag{sbm}{Sensing}
\cvtag{sbm}{Observation} \newline
Extends from SSN
\cvtag{ssn}{SensingDevice}
\cvtag{ssn}{Observation}
\cvtag{ssn}{Property}
\cvtag{ssn}{FeatureOfInterest}

& The SBMS ontology represents information available for building operational analysis. It is an extension of SSN ontology introducing the concept of data points \\

A semantics-driven framework to enable demand flexibility control applications~\cite{de2025semantics} & Zones, Spaces, Thermostats, Occupancy sensors, Temperature sensors, Temperature setpoints, minimum and maximum temperature setpoints, and grid signals & Uses Brick and SAREF
\cvtag{bri}{Zone}
\cvtag{bri}{Space}
\cvtag{bri}{Occupancy Sensor}
\cvtag{bri}{Thermostat}
\cvtag{bri}{Max Air Temperature Setpoint}
\cvtag{bri}{Temperature Setpoint}
\cvtag{sar}{Sensor}
\cvtag{sar}{Property}
\cvtag{sar}{Actuator}
\cvtag{s4b}{UnitaryControlElement}
\cvtag{s4b}{BuildingSpace} & Integration of multi contextual building data through Brick and SAREF to support control applications \\

Smart building diagnosis\newline ~\cite{ploennigs2014adapting} & Process, Observation, Properties, Environment & New classes using namespace PHY

\cvtag{phy}{PhysicalProcess}
\cvtag{phy}{FeatureLink}
\cvtag{phy}{Anomaly}
\cvtag{phy}{Property}
\cvtag{phy}{Cause}
 \newline
Extends from SSN
\cvtag{ssn}{FeatureOfInterest}
\cvtag{ssn}{Property}
\cvtag{ssn}{Stimulus}
\cvtag{ssn}{Sensor}
\cvtag{ssn}{Observation}
& This custom ontology utilised the SSN ontology and extended it to model physical and cause-effect relationships between sensors to support building applications to automatically
derive complex models for analytics tasks such as prediction and diagnostics. \\

\SetCell[c=4]{c, bg=s4b, font=\footnotesize\bfseries} Multiple-ontology use case scenarios & & & \\

Energy Analysis~\cite{zhang2021linking} & Building Basic Information, Geometry and Topology, Energy Consumption, Environment Condition & Uses BOT, SOSA, and a custom KPI ontology
\cvtag{kpi}{Input}
\cvtag{kpi}{Result}
\cvtag{kpi}{Formula}
\cvtag{kpi}{Parameter}
\cvtag{kpi}{Operator}
\cvtag{bot}{Building}
\cvtag{bot}{Space}
\cvtag{bot}{Zone}
\cvtag{sos}{Platform}
\cvtag{sos}{Sensor}
\cvtag{sos}{Observation}
\cvtag{sos}{FeatureOfInterest}
\cvtag{sos}{Property}
& This linked data model was used to connect data silos to evaluate building's energy consumption efficiency \\

Semantic modeling of building operations~\cite{delgoshaei2022semantic} & Equipment and Systems, Sensing and Actuation, Building Topology, Unit Of Measurement, Temporal Information & Uses SAREF4BLDG, SOSA, BOT, and Brick
\cvtag{bot}{adjacentZone}
\cvtag{bri}{HVAC Zone}
\cvtag{bri}{Supply Air Temperature Sensor}
\cvtag{bri}{VAV}
\cvtag{bri}{AHU}
\cvtag{bri}{Cooling Valve}
\cvtag{bri}{Mixed Air Temperature Sensor}
\cvtag{bri}{Exhaust Fan}
\cvtag{sos}{madeObservation}
\cvtag{sos}{resultTime}
\cvtag{s4b}{BuildingSpace}
& It emphasised on the semantic representation of building assets with inference mechanisms for building applications ranging from fault detection and diagnostics, asset management and maintenance and context-aware control \\

Data integration for delivering data-driven application~\cite{mavrokapnidis2021linked} & Dynamic data, Sensor tags, Building topology, Systems equipment, Building elements, Product properties & Uses BOT, Brick, BEO, and PROPS
\cvtag{bot}{adjacentElement}
\cvtag{bot}{Space}
\cvtag{bri}{Thermostat}
\cvtag{bri}{Space}
\cvtag{bri}{Zone Air Temperature Set Point}
\cvtag{bri}{timeseries}
\cvtag{oth}{Window}

& The linked data was adopted for data integration between multiple diverse data sources to facilitate the development of digital twin applications \\

Ontology-based integration for thermal zoning analysis~\cite{wu2023ontology} & Hvac system, Internal heat gain, Building, Weather & Uses Brick, BOT, and BES
\cvtag{bri}{Location}
\cvtag{bri}{AHU}
\cvtag{bri}{VAV}
\cvtag{bri}{Cooling Temperature Setpoint}
\cvtag{bri}{Heating Temperature Setpoint}
\cvtag{bri}{Occupancy}
\cvtag{bri}{Equipment}
\cvtag{bot}{Element}
\cvtag{bot}{Space}
\cvtag{oth}{Weather}
& Semantic integration of multifaceted building data for automatic building energy modeling and thermal zoning
\\

\SetCell[c=4]{c, bg=s4b, font=\footnotesize\bfseries} Comparative analyses of semantic models & & & \\

Ontologies for Building Energy Applications\newline~\cite{pritoni2021metadata} & Zones and Spaces, Envelope, Building System and Equipment, Control Devices, Sensors and Actuators & Uses BOT, Brick, SSN/SOSA, SAREF, and REC
\cvtag{bri}{VAV}
\cvtag{oth}{Device}
\cvtag{s4b}{Flow Terminal}
\cvtag{ssn}{System}
\cvtag{s4b}{BuildingObject}
\cvtag{oth}{VirtualBuildingComponent}
\cvtag{bri}{HVAC Zone}
\cvtag{bri}{Room}
\cvtag{bot}{Space}
\cvtag{oth}{Office}

& Comparative analysis of five distinct ontologies for modeling an office building to support tasks such as energy audits, AFDD, and optimal control
\\

Brick for portable smart building applications~\newline \cite{balaji2018brick} & Sensor data points, Building system and equipment, spatial contexts, building controllable resources (air and water) & Uses Brick and Project Haystack\newline
\cvtag{bri}{VAV}
\cvtag{bri}{Room}
\cvtag{bri}{AHU}
\cvtag{bri}{HVAC Zone}
\cvtag{bri}{Zone Temperature Sensor}
\cvtag{bri}{Supply Air Flow Sensor}
\cvtag{oth}{zone}
\cvtag{oth}{temp}
\cvtag{oth}{sensor}
\cvtag{oth}{ahuRef}
\cvtag{oth}{equipRef}
\cvtag{oth}{area}
\cvtag{oth}{siteRef}
\cvtag{oth}{vav} & Comparative evaluation of Brick and Project Haystack to support a robust semantic layer for portable and scalable smart building applications
\\


\hfill\emph{Legend:} & \SetCell[c=3]{l}
\begin{tabular}[t]{@{}l@{}}
\cvtag{sar}{saref:}\cvtag{s4b}{s4bldg:}\cvtag{s4s}{s4syst:}\cvtag{icb}{icbms:}\cvtag{bot}{bot:}\cvtag{sos}{sosa:}\cvtag{kpi}{kpi:}\cvtag{bri}{brick:}
\\[2pt]
\cvtag{sbm}{sbms:}\cvtag{ssn}{ssn:}\cvtag{phy}{phy:}\cvtag{oth}{other}
\end{tabular}\\

\cline[2pt,white]{-}
\end{longtblr}

\begin{longtblr}[
  caption = {Usecase Quantitative Analysis.},
  label = {tab:quantitative}
]{
  colspec = {Q[0.60,l,m] Q[1.0,l,m] Q[0.70,l,m] Q[0.8,l,m] Q[1.3,l,m] Q[1.0,l,m] Q[1.0,l,m]},
  rows    = {valign=t},
  cells   = {font=\footnotesize\selectfont},
  rowhead = 1,
  hlines
}
\cline[2pt]{-}
\textbf{Use case} \newline \textbf{and OEC rating} & \textbf{No. of Concepts} \newline \textbf{(total classes; \(P\)=parent classes)} & \textbf{Used Concepts} & \textbf{No. of concepts created} & \textbf{OIR} \newline \textbf{(parent-class basis)} & \textbf{NTE}& \textbf{Method of concept coverage}\\
\SetCell[c=7]{c, bg=s4b, font=\footnotesize\bfseries} Single-ontology use case scenarios & & & & & & \\
~\cite{bjornskov2023ontology}\newline \cvtag{oeccomplete}{complete} & Sarefcore: 29; \(P=20\) \newline
Saref4bldg: 65; \(P=8\) \newline
Saref4syst: 3; \(P=3\) & 22 & 8 & 5/20 = 25\% \newline 4/8 = 50\% \newline 3/3 = 100\% & 8/30 = 26.67\% & Ontology class specialisation\\
~\cite{van2020validating}\newline \cvtag{oeccomplete}{complete} & Sarefcore: 29; \(P=20\) & 10 & 2 & 9/20 = 45\% & 2/12 = 16.67\% & Ontology external inheritance\\

~\cite{weerdt2021making} \cvtag{oeccomplete}{complete} & Sarefcore: 29; \(P=20\) & 8& 1& 6/20 = 30\%& 1/9 = 11.1\% & Ontology external inheritance\\
~\cite{van2024officegraph} \cvtag{oecpartial}{partial} & Sarefcore: 29; \(P=20\) \newline Saref4bldg: 65; \(P=8\) \newline Saref4ener: 76; \(P=47\) & 15 & 9 & Sarefcore: 3/20~=~15.0\% \newline Saref4bldg: 1/8~=~12.5\% \newline Saref4ener: 1/47~=~2.12\% & 9/24~=~37.5\% & Application ontology extension
\\
~\cite{kirnapci2025ontology} \cvtag{oecpartial}{partial} & Sarefcore: 29; \(P=20\) & 8 & 4 & 6/20~=~30\% & 4/12~=~33.33\% &  Ontology external inheritance
\\
~\cite{xu2024semantic}\newline \cvtag{oecpartial}{partial} & Brick: 1450; \(P=5\) & 5 & 0 & 3/5~=~60\% & 0\% & N/A\\

~\cite{chiosa2024portable}\newline \cvtag{oecpartial}{partial} &Brick: 1450; \(P=5\) & 6 & 0 & 3/5~=~60\% & 0\% & N/A\\

~\cite{zheng2023towards}\newline \cvtag{oecpartial}{partial} & Brick: 1450; \(P=5\) & 11 & 8 & 3/5~=~60\% & 8/19~=~42.10\% & Ontology class specialisation
\\

~\cite{sun2022enabling}\newline \cvtag{oeccomplete}{complete} & Brick: 1450; \(P=5\)& 5& 0 & 3/5 = 60\% & 0\% & N/A\\

~\cite{hwang2025dt}\newline \cvtag{oecpartial}{partial} & Brick: 1450; \(P=5\)& 5& 0 & 3/5 = 60\% & 0\% & N/A\\
~\cite{fierro2020shepherding}\newline \cvtag{oecpartial}{partial} & Brick: 1450; \(P=5\)& 3 & 0 & 4/5 = 80\% & 0\% & N/A\\

\SetCell[c=7]{c, bg=s4b, font=\footnotesize\bfseries} Two-ontology use case scenarios & & & & & & \\
~\cite{ji2017intelligent}\newline \cvtag{oeccomplete}{complete} & ICBMS extends SAREF \newline  ICBMS: Undisclosed \newline Sarefcore: 29; \(P=20\) \newline Saref4bldg: 65; \(P=8\)& 4& 17& Sarefcore: 2/20~=~10\% \newline Saref4bldg: 1/8~=~12.5\% & 17/21 = 80.95\% & Ontology reuse and \newline Ontology external inheritance\\

~\cite{kuvcera2018semantic}\newline \cvtag{oeccomplete}{complete}&SBMS extends SSN ontology \newline
SBMS: 47; \(P=18\) \newline
SSN: 21; \(P=21\)& 5 & 9& SSN: 4/21 = 19.05\% \newline SBMS: 10/18 = 55.55\% & 9/14 = 64.28\% & Ontology reuse and \newline Ontology external inheritance \\

~\cite{de2025semantics}\newline \cvtag{oeccomplete}
{complete} & Brick: 1450 \newline
Saref4bldg: 65; \(P=8\) \newline
Sarefcore: 29; \(P=20\) & 63 & 5 & Sarefcore: 3/20 = 15\% \newline Saref4bldg: 4/8 = 50\% \newline Brick: 3/5 = 60\% & 5/68 = 7.35\% & Ontology integration and \newline Ontology external inheritance\\

~\cite{ploennigs2014adapting}\newline \cvtag{oecpartial}{partial} & SB ontology extends SSN ontology \newline SB ontology: 122; \(P=7\) \newline
SSN ontology: 21; \(P=21\) & 2 & 19 & SSN: 5/21 = 23.81\% \newline
SB ontology: 6/7 = 85.71\% & 19/21 = 90.48\% & Ontology reuse and \newline Ontology class specialisation\\

\SetCell[c=7]{c, bg=s4b, font=\footnotesize\bfseries} Multiple-ontology use case scenarios & & & & & & \\
~\cite{zhang2021linking}\newline \cvtag{oeccomplete}{complete} & BOT: 8; \(P=4\)\newline
SOSA: 16; \(P=16\) \newline
Custom KPI: 21; \(P=21\) &  BOT: 2 \newline
SOSA: 5 \newline
KPI: 5 & BOT: 10 \newline
SOSA: 7 \newline KPI: 7 & BOT: 2/4~=~50\% \newline SOSA: 5/16~=~31.25\% \newline KPI: 5/21~=~23.81\%
&  N/A & Ontology integration\\
~\cite{delgoshaei2022semantic}\newline \cvtag{oecpartial}{partial}& BOT: 8; \(P=4\) \newline Saref4bldg: 65; \(P=8\) \newline SOSA: 16; \(P=16\) \newline Brick: 1450; \(P=5\) & BOT: 1 \newline Saref4bldg: 1 \newline SOSA: 3 \newline BRICK: 7& BOT: 11 \newline Saref4bldg: 11 \newline SOSA: 9 \newline BRICK: 5& BOT: 1/4~=~25\% \newline Saref4bldg: 1/8~=~12.5\% \newline SOSA: 2/16~=~12.5\% \newline Brick: 3/5~=~60\% & N/A & Ontology integration\\

~\cite{mavrokapnidis2021linked}\newline \cvtag{oecpartial}{partial}& BOT: 8; \(P=4\) \newline Brick: 1450; \(P=5\) \newline BEO: 186; \(P=4\) & BOT: 4 \newline Brick: 3 \newline BEO: 3& BOT: 6 \newline Brick: 7 \newline BEO:7& BOT : 1/4~=~25\% \newline
Brick: 3/5~=~60\% \newline BEO: 1/4~=~25\%& N/A & Ontology integration\\

~\cite{wu2023ontology}\newline \cvtag{oeccomplete}{complete} & BOT: 8; \(P=4\) \newline Brick:1450; \(P=5\) \newline BES: 11; \(P=3\) & BOT: 5, \newline Brick: 5, \newline BES: 7& BOT: 12 \newline
Brick: 12 \newline BES: 10 & BOT: 2/4~=~50\% \newline Brick: 3/5~=~60\% \newline BES: 7/11~=~63.64\% & N/A & Ontology integration \\

\SetCell[c=7]{c, bg=s4b, font=\footnotesize\bfseries} Comparative analyses of semantic models & & & & & & \\

~\cite{pritoni2021metadata}\newline \cvtag{oeccomplete}{complete} & BOT: 8; \(P=4\) \newline Brick: 1450; \(P=5\) \newline SSN: 21; \(P=21\) \newline Saref4bldg: 65; \(P=8\) \newline REC: 178; \(P=8\) & BOT: 1 \newline  Saref4bldg: 3 \newline  Brick: 3 \newline SSN:1 \newline REC: 4 & BOT: 2 \newline Saref4bldg: 0 \newline BRICK: 1 \newline SSN: 3 \newline REC: 1 & BOT: 1/4~=~25\% \newline Brick: 2/5~=~40\% \newline SSN: 1/21~=~4.76\% \newline Saref4bldg: 2/8~=~25\% \newline REC: 2/8~=~25\% & N/A & Ontology integration\\

~\cite{balaji2018brick}\newline
Brick: \cvtag{oeccomplete}{complete} \newline
PH: \cvtag{oecpartial}{partial} &Brick: 1450; \(P=5\) \newline PH: 514 tags & Brick: 5 \newline PH: 8 & N/A & Brick: 3/5 = 60\% \newline PH: 8/514 = 1.55\% & 0\% & N/A
\\

\hfill\emph{Legend:} & \SetCell[c=3]{r}\cvtag{oeccomplete}{OEC: complete}\cvtag{oecpartial}{OEC: partial}
\\

\cline[2pt,white]{-}
\end{longtblr}

{\footnotesize\emph{Table note:} The coloured labels in the first column report the OEC rating for each use case. The \emph{No. of Concepts} column reports total ontology classes or tags. \(P\) reports the parent-class denominator used for the OIR; therefore, rates such as \(5/20\) are computed from parent classes, not from the full class count. In integration rows, \emph{Number of Concepts Created} is ontology-specific and denotes concepts not directly covered by the listed ontology but addressed through the other ontology or ontologies in the integrated model.\par}

A fault diagnosis use case combined PH and Brick to improve the portability of an AHU diagnostic tool for ventilation units with heat recovery systems~\cite{nehasil2021versatile}. The application used performance assessment rules and expert-assigned metadata tags to characterise measured quantities, locations, and control roles across AHU configurations. Neither PH nor Brick alone was sufficient for the required semantic data, motivating a unified ontology. OEC is None because sufficient class-level mappings were unavailable for quantitative concept-to-class analysis; the study was therefore omitted from Table~\ref{tab:usecase}, although it still shows how combined semantic models can support portable AHU fault detection.

These two-ontology use cases show that pairing semantic models can improve task-specific coverage when the models provide complementary semantic support. They also indicate that two-model strategies are most useful when the application spans concepts that are not adequately represented by a single model, but their interpretability still depends on explicit concept-to-class mappings and clearly documented ontology use.

\subsection{Multiple-ontology use case scenarios}

This section examines use cases in which multiple ontologies are employed to represent the data required by building operational applications. This approach typically combines complementary ontologies to extend the descriptive capacity of a single model when it cannot adequately capture all relevant building concepts or data elements. The following descriptions analyse studies that adopted this multiple-ontology strategy to support specific operational applications.

A KPI calculation use case used linked data to integrate BIM and sensor network data for building performance benchmarking and energy evaluation~\cite{zhang2021linking}.
The demonstrated application automated the retrieval and processing of building and sensing data to compute performance indicators, illustrated through electricity consumption per area per day in real buildings.
The required concepts covered building identity, geometry and topology, spatial hierarchy, building elements, sensing entities, observations, resource consumption, environmental variables, and KPI-specific elements such as inputs, outputs, formulae, parameters, operators, time periods, and calculation results.
OEC is Complete because the paper explicitly documents the ontology structures and their roles in the implemented framework. BOT, SOSA, and a newly developed KPI ontology were used to distribute the semantic representation across complementary models.
The OIRs were 25\% for BOT, 31.25\% for SOSA, and 23.81\% for the KPI ontology. Since the required concepts were covered through this integrated modelling strategy, no NTE was identified, and ontology integration constituted the method of concept coverage.

A related study reused pre-existing KPI ontologies to support automated and adaptable KPI computation in BEMS using real data processing and interoperable configuration~\cite{schmid2025continuous}. However, OEC is None because the semantic layer is not described in sufficient detail and no explicit concept-to-class mappings are provided for the KPI computation concepts.

A semantic infrastructure for building system operations used multiple ontologies to support knowledge representation and reasoning across applications such as control, automation, analytics, compliance checking, maintenance, and integration~\cite{delgoshaei2022semantic}.

The framework combined Brick, SOSA, BOT, SAREF4BLDG, QUDT, and the Time Ontology~\cite{timeOntology} to support operational scenarios such as diagnostics, reasoning, asset management, maintenance, and context-aware control. A custom namespace was introduced for concepts not covered by these models, especially geometry-related and time-series equipment properties.
OEC is Partial because the paper demonstrates ontology use through rules, queries, and illustrative application cases, but does not provide a complete concept-to-class mapping for all represented concepts across the full framework.
The OIRs were 25\% for BOT, 12.5\% for SAREF4BLDG, 12.5\% for SOSA, and 60\% for Brick. These values show that the framework relied on the combined contribution of multiple ontologies, making ontology integration the main concept coverage strategy.

A linked data paradigm connected static design information with dynamic operational data to support building Digital Twin applications~\cite{mavrokapnidis2021linked}.
Rather than addressing a specific control task, the use case demonstrated unified querying across heterogeneous building and operational data sources. The represented concepts covered topology, envelope, thermal properties, HVAC and sensing devices, operational data references, and temperature observations.
IFC was used as the original design data source, but its complexity, non-modular structure, and rigid schema form~\cite{pauwels2017semantic} motivated its transformation into BOT using the IFC-to-LBD converter framework~\cite{bonduel2018ifc}. BOT was then combined with Brick for operational entities, the Building Element Ontology (BEO)~\cite{BEO} for building element types, and the Building-Related Properties (PROPS) ontology~\footnote{\url{https://github.com/maximelefrancois86/props}} for properties specific to building elements or equipment.
BEO and PROPS were not discussed earlier because their primary focus is on the design and construction phases. OEC is Partial because the paper presents the integration structure and illustrates class usage through the conceptual model and query workflow, but does not provide a complete concept-to-class mapping for the full set of represented concepts.
The OIRs were 25\% for BOT, 60\% for Brick, and 25\% for BEO. The semantic requirements of this use case were therefore addressed through ontology integration.

Thermal zoning analysis provides another multiple-ontology example, where data from BIM, BMS, weather, and occupancy sources were integrated to support rule-based reasoning for thermal load simulation and predictive control~\cite{wu2023ontology}.
The required concepts covered spatial entities, HVAC components, thermal control parameters, occupancy information, and weather-related data.
OEC is Complete because the study provides explicit concept-to-class mappings for both HVAC system data and weather data. Brick, BOT, and a weather information ontology were used in the case study.
The proposed weather information ontology, identified by the prefix BES, is not publicly available, although the study states that it consists of three main classes. The OIRs were 60\% for Brick, 50\% for BOT, and 63.64\% for the BES ontology.
Together, the three ontologies provided the semantic coverage required for the application through ontology integration.

In multiple-ontology scenarios, broader BEM applications often require several semantic models to be coordinated within a single framework. The main insight is that multi-model strategies can provide wider semantic coverage, but they also increase the need for transparent alignment, explicit concept-to-class mappings, and reusable modelling patterns.

\subsection{Comparative analyses of semantic models in prior studies}
In addition to the application-oriented studies reviewed in this survey, some prior works explicitly compared semantic models for building-related domains. These studies help contextualise the use case findings by examining where semantic models overlap, complement one another, and differ in their strengths and limitations in relation to specific building applications.

A comparative review assessed the suitability of selected semantic models for building energy applications, including energy audits, automated fault detection and diagnostics, and optimal control of HVAC systems~\cite{pritoni2021metadata}.
The comparison examined how spatial entities, envelope components, building systems and equipment, control entities, and sensing and actuation entities were represented, together with their associated properties.
BOT, Brick, SSN, SAREF4BLDG, and RealEstateCore were analysed against these requirements, showing both overlapping and complementary coverage across the targeted applications.
OEC is Complete because the study provides sufficient concept-level evidence to inspect the coverage of the compared models. The OIRs were 25\% for BOT, 40\% for Brick, 4.76\% for SSN, 25\% for SAREF4BLDG, and 25\% for REC.
Despite this coverage, the study identified persistent limitations in geometry, envelope properties, equipment properties, units of measurement, and control strategy concepts such as schedules and setpoints.

A comparative evaluation examined Project Haystack, IFC, and Semantic Sensor Web frameworks for portable smart building applications~\cite{bhattacharya2015short}. The required concepts covered equipment, sensing and actuation points, spatial entities, control relationships, measurement properties, operational state, and application-specific metadata. The study found partial and uneven support across the three schemas: Haystack better covered common HVAC metadata, IFC provided richer spatial and asset modelling, and Semantic Sensor Web remained more generic and fragmented. Evidence from three commercial buildings and 87 applications highlighted incomplete metadata coverage, limited application-relevant connections, weak uncertainty modelling, and limited flexibility for novel sensors and subsystems.
OEC is None because the study did not explicitly map the required concepts to ontology classes in a way that supports concept-to-class traceability; therefore, it is excluded from Table~\ref{tab:usecase}.

Building on the earlier comparative analysis, Brick was introduced as a uniform metadata schema for portable smart building applications~\cite{balaji2018brick}.
The work responded directly to limitations observed in Project Haystack, IFC, and Semantic Sensor Web frameworks, particularly incomplete metadata coverage, limited operational consistency, and insufficient support for portable application queries.
Brick retained the practical tagging intuition of Project Haystack but formalised it through a class hierarchy and structured modelling patterns for building equipment, points, spatial contexts, and controllable resources such as air and water.

OEC is Complete for Brick because the paper documents the ontology hierarchy, core modelling structure, and queryable application mappings used to represent the targeted concepts. Project Haystack is classified as Partial because the manuscript reports relevant tags, but its tag-based representation provides less traceable class-level structure for the analysed concepts.
The OIR for Brick was 60\%, whereas Project Haystack achieved 1.55\%, based on the tags explicitly reported in the manuscript. Through validation across six real buildings and several portable applications, the study showed how Brick addressed some of the shortcomings identified in the earlier comparative analysis.

A comparative review assessed the suitability of eight ontologies for semantic interoperability in demand-side management~\cite{de2022towards}. 
The ontologies were analysed against key concept groups, including spatial information, building energy systems, control and topology, measurement setup, measurable properties, and grid interactivity.

The comparison was grounded in metadata quality metrics~\cite{ochoa2009automatic}, with each ontology evaluated according to its ability to represent these concepts and support context-aware demand-side management.
The results showed that none of the selected ontologies fully represented all identified concepts, although Brick and SAREF were reported as the most suitable models for covering much of the data required by demand-side management applications.
The study further argued that ontology-based demand-side management architectures can support energy flexibility, operational efficiency, cost performance, and environmental comfort through intelligent and responsive control strategies.
OEC is None because the study did not specify the classes or tags used to represent the identified concepts; therefore, it was excluded from Table~\ref{tab:usecase}.

The use cases reviewed in this section show that ontology adoption can support diverse BEM applications, including monitoring, simulation, energy analysis, fault detection, control, optimisation, and performance assessment. In relation to \emph{RQ3}, the central finding is that ontology use in operational BEM is task-dependent. Single ontologies are effective when the application remains close to their original modelling scope, whereas two-ontology and multiple-ontology strategies are used when broader contextual, computational, temporal, or cross-domain representation is required. The analysis also shows that semantic coverage is more consistently documented for physical building entities, devices, sensors, observations, equipment, and operational metadata than for computational, control, grid-interactive, and decision-support concepts. Practical semantic coverage therefore often depends on ontology reuse, integration, class specialisation, external inheritance, or application-specific extension. These findings lead directly to \emph{RQ4} by showing that the main limitation is not only ontology availability, but also how existing semantic models are selected, combined, extended, and documented in operational BEM applications.

\section{Discussion on identified limitations and forthcoming steps}
\label{sec:Discussion}

Our semantic model review and use case analysis reveal that the main challenge for operational BEM is not the absence of ontologies, but the inconsistent ways in which existing ontologies represent, combine, and extend operational concepts.
This discussion therefore focuses on recurring limitations in semantic coverage, extension practice, interoperability, and the representation of computational and decision-support concepts.

This section is grounded in the ontology catalogue presented in Table~\ref{tab:other-ontology}, the use case scenarios summarised in Table~\ref{tab:usecase}, and the quantitative synthesis reported in Table~\ref{tab:quantitative}. Together, these results show how semantic models are selected, combined, extended, and used in operational BEM studies, providing the basis for identifying the limitations and forthcoming steps associated with \emph{RQ4}.

\subsection{Identified limitations}

\paragraph{\textbf{Ontology instantiation and extension patterns}}
Table~\ref{tab:quantitative} shows that use cases with many missing concepts often require ontology reuse or integration, whereas smaller gaps are usually addressed through class specialisation or external inheritance. Ontology extension is therefore a recurring feature of applied BEM semantic modelling, especially when operational tasks involve cross-domain requirements or concepts beyond the scope of a single ontology.

\paragraph{\textbf{Semantic adequacy for operational BEM}}
Operational BEM requires semantic models that represent physical, environmental, occupancy, computational, and control-related concepts at the level of detail required by the task. The reviewed use cases show that interoperability depends not only on vocabulary alignment, but also on whether the selected ontology or ontology combination can explicitly represent the semantic requirements of a concrete application.

\paragraph{\textbf{Spatial, device, and system representation}}
The representation of zones and spaces illustrates a recurring trade-off between flexibility and operational specificity. BOT and SAREF4BLDG provide flexible spatial representations, but their broad treatment of zones and spaces can lead to divergent interpretations. Brick distinguishes zones from spaces more explicitly by defining zones as collections of spaces for controlled operations, but its predefined categories can be less adaptable when zones are defined within a single space.

Sensors, actuators, and building systems are supported by SSN/SOSA, SAREFcore, SAREF4BLDG, Brick, and RealEstateCore, but differences in modelling practice can affect interoperability. For example, SAREF-based studies may represent a temperature sensor either as \texttt{saref:TemperatureSensor} or through a combination of \texttt{saref:Device} and \texttt{saref:Sensor}~\cite{bjornskov2023ontology, van2020validating}. Brick offers more structured building-specific classes, whereas SSN/SOSA and SAREF provide broader flexibility; however, this flexibility requires clearer philosophical and technical guidance for modelling building entities and their requirements~\cite{hull1987semantic}. This evidence therefore suggests that explicit concept-to-class mappings are often more important for interoperability than the number of ontology classes used.

HVAC systems are major energy consumers in buildings, but their semantic representation remains uneven. SSN/SOSA, SAREF, and RealEstateCore provide general HVAC representations, whereas Brick offers more detailed equipment hierarchies and relationships. However, the use cases show that even Brick is not complete on its own, since broader geometric, control, temporal, and grid-related requirements often require additional modelling support.

\paragraph{\textbf{Occupancy and intelligent control}}
Occupancy is central to BEM because it affects building performance and energy efficiency. As defined in Section~\ref{sec:core concepts}, it includes occupant behaviour, occupancy count, and occupant detection. The OP ontology models occupant detection for energy-related building management, while the extended Brick ontology captures both occupancy behaviour and count~\cite{luo2022extending, opontology}. However, both provide weaker support for the computational processes needed to infer building status from passive occupancy data. This reflects a broader use-case pattern: current semantic models represent physical entities and monitored variables more consistently than the operational intelligence needed to interpret and act on them.

Intelligent control systems regulate energy use, environmental conditions, and occupant satisfaction by responding to factors such as occupancy, weather, and resource availability. Semantic frameworks can link control intelligence with building components, improving consistency, scalability, and adaptability in smart-building applications~\cite{teixeira2025semantic}. However, the reviewed use cases show that semantic support remains uneven for schedules, event logic, optimisation variables, computational requirements, and control relationships.

\paragraph{\textbf{Computational and decision-support concepts}}
Computational and decision-support concepts remain weakly integrated in current BEM ontologies. Tasks such as \emph{KPI} evaluation, performance \emph{assessments}, and \emph{service} execution are essential for adaptive operation, but they are often modelled separately rather than as integrated parts of an operational workflow. Existing KPI-oriented ontologies provide useful structures for performance tracking~\cite{li2019enhancing,kpiOntology}, yet they do not fully link KPI computation, building assessment, and responsive service implementation. The use case analysis therefore indicates a coverage gap in representing the abstract, dynamic, and decision-support concepts required for intelligent building operation.

\paragraph{\textbf{Energy flexibility and grid interaction}}
The European Union's emphasis on building energy flexibility for decarbonisation requires effective communication and coordination among distributed energy resources and stakeholders~\cite{flexibility}. Semantic modelling can support this by standardising data representation across flexibility-related sources, but demand response and broader flexibility concepts remain insufficiently investigated. Existing ontologies only partially cover the diverse resources identified in the literature~\cite{li2021energy}, and the use case analysis shows that no single ontology fully supports the semantic requirements of grid-interactive applications.

\subsection{Forthcoming steps}

Building on these limitations, the following directions indicate where ontology development, comparison, and application practice can be strengthened for operational BEM.

\paragraph{\textbf{SAREF and Brick in operational BEM}}
One immediate direction is the systematic comparison of SAREF and Brick in operational BEM. Because their use may reflect not only concept coverage but also standardisation context, tooling, project requirements, and community familiarity, future work should evaluate them across common use cases using transparent concept-to-class mappings.

\paragraph{\textbf{AI agents and semantic operational intelligence}}
In relation to computational and decision-support concepts, recent advances in natural language processing and AI agents create opportunities for intelligent BEM applications~\cite{yoon2025ontology, devmane2026ontosage}. When integrated with semantic layers, AI agents can support building knowledge queries, operational-state interpretation, and BEM tasks. However, their reliability depends on explicit and traceable representations of task-relevant concepts, class mappings, control relationships, computational workflows, and data provenance. Semantic models therefore define not only what AI agents can access, but also what they can interpret, explain, and act upon in building operation.

\paragraph{\textbf{Towards coordinated ontology development}}
These limitations raise a key question: should operational BEM rely on a unified ontology or on modular ontologies linked for specific tasks? A unified model could improve consistency, similar to how the Unified Medical Language System integrates biomedical vocabularies~\cite{bodenreider2004unified}. A modular approach may better support distinct subdomains such as geometry, HVAC, occupancy, environmental conditions, control, and computational processes.

The reviewed use cases suggest that practical interoperability in BEM is increasingly multi-ontology in nature; therefore, the central challenge is how ontologies can be aligned, extended, and combined in a principled and sustainable way. This is particularly relevant for energy flexibility and grid-interactive applications, where future frameworks should support consistent resource representation, robust ontology extension, and systematic integration of new flexibility concepts without undermining semantic coherence.

Governance requirements differ between these approaches. A unified ontology would require a standards body or industry consortium to manage updates, conflicts, compatibility, and versioning, whereas modular ontologies shift responsibility toward integration, shared modelling principles, and semantic alignment. The evidence from this survey suggests that future progress in BEM semantic modelling will depend less on inventing isolated new ontologies and more on improving explicit modelling guidance, extension mechanisms, ontology alignment, and the representation of computational and operational processes.

Table~\ref{tab:other-ontology} shows that many ontologies remain underutilised in real-world applications, partly because of limited openness, accessibility, or maintenance. Fragmentation is another concern, as researchers may create extensions or new ontologies when existing models lack required entities, contributing to overlapping models with slight variations. A coordinated strategy centred on collaboration, reuse, explicit concept coverage, and structured integration could reduce this proliferation and improve how ontologies are documented, extended, aligned, and applied in real building operational scenarios.

\section{Conclusions}
\label{sec:conclusion-future}

This survey examined semantic modelling technologies for IoT-based BEM, focusing on the building operational phase. It reviewed sixty-one semantic models and analysed their application in more than twenty concrete BEM use cases through one, two, and multiple-ontology strategies. It also introduced OEC as an evidence-oriented perspective for assessing whether task-relevant operational concepts are traceably mapped to ontology classes in applied studies.
The survey answers why ontologies are needed for operational BEM, which models are used, how they are applied in the use cases, and what limitations remain.

The review shows that semantic models are increasingly important and needed for interoperability, data integration, reusable querying, and context-aware BEM applications.
The quantitative analysis further shows that applied studies frequently address missing concepts through ontology integration, reuse, class specialisation, external inheritance, or application-specific extensions. While these strategies improve coverage for individual applications, they also raise concerns about consistency, reusability, and long-term semantic interoperability. Future work should therefore prioritise principled extension practices, clearer modelling guidance, and improved alignment between existing semantic models, rather than continuing to develop isolated ontologies with overlapping scope and purpose.
At the same time, the use case analysis confirms that no single ontology provides complete coverage of the semantic requirements of building operation. Existing models are comparatively mature in representing physical building assets, systems, devices, and measurements, but less consistent in representing the abstract operational concepts needed for decision support, including performance evaluation, control logic, computational workflows, and energy flexibility coordination.

These findings are based on the evidence available in the reviewed publications and associated resources. Where semantic evidence was incomplete or unavailable, quantitative analysis was limited; the results should therefore be read as evidence-based indicators of reported ontology use rather than exhaustive measures of each ontology's modelling capacity.

Overall, semantic modelling has substantial potential to support more intelligent, efficient, and sustainable building operation. Realising this potential requires semantic representations that go beyond describing building assets to also capture the computational and decision-support processes through which BEM systems interpret data, evaluate performance, and trigger actions. This direction is essential for developing more interoperable, generalisable, and context-aware BEM systems.

\section*{Statements and Declarations}

\paragraph{Funding}
This work was partially supported by the European Commission through the SATO project, ref. H2020-IA-957128, and SMART2B project, ref. H2020-IA-101023666, and Funda\c{c}\~{a}o para a Ci\^{e}ncia e Tecnologia (FCT) through the LASIGE Research Unit, ref.
UID/00408/2025, DOI \url{https://doi.org/10.54499/UID/00408/2025}, and by the Unite!Widening Ph.D. grant 13/BD/2025.

\paragraph{Competing interests}
The authors have no competing interests to declare that are relevant to the content of this article.

\paragraph{Author contributions}
\textbf{M. A.}: Conceptualisation, Methodology, Validation, Investigation, Data curation, Writing – original draft, Writing – review \& editing, Visualisation. 
\textbf{V. C.}: Conceptualisation, Methodology, Writing – review \& editing, Visualisation, Supervision. 
\textbf{P. F.}: Conceptualisation, Methodology, Resources, Writing – review \& editing, Supervision, Project administration, Funding acquisition.

\paragraph{Data availability}
No new datasets were generated or analysed during the current study. The reviewed literature and ontology resources are cited in the manuscript.

 \bibliographystyle{elsarticle-num}
 \bibliography{mainref}
\clearpage

\appendix

\section{Automated screening keyword groups}
\label{app:screening_rules}

The automated screening stages used normalised title, abstract, and keyword metadata. The text was normalised for punctuation, capitalisation, spacing, and character-formatting differences before screening. British and American spelling variants were included where relevant to avoid excluding records because of spelling differences.
The keyword groups used in the scope-based and rule-based screening were:
\begin{itemize}
    \item \textbf{Building and energy domain:} building, buildings, smart building, building system, building automation, BEM, BEMS, HVAC, energy management, energy efficiency, building operation, occupancy, thermal comfort, demand response, smart home, and smart grid.
    \item \textbf{Semantic modelling:} ontology, ontologies, semantic, semantics, linked data, RDF, OWL, knowledge graph, metadata, data model, information model, schema, vocabulary, and taxonomy.
    \item \textbf{Model-oriented:} model, modelling, modeling, framework, architecture, representation, interoperability, and integration.
    \item \textbf{Review-oriented:} review, survey, systematic review, state of the art, comparison, and comparative.
    \item \textbf{Application oriented signals:} optimisation, optimization, management, analytics, operation, monitoring, benchmarking, control, fault detection, diagnosis, and service.
\end{itemize}
The title-screening rule can be summarised as:

\[
\begin{aligned}
\text{retain} ={}& (\text{building or energy domain}) \\
&{}\land (\text{semantic modelling} \\
&{}\lor \text{model oriented} \\
&{}\lor \text{review oriented} \\
&{}\lor \text{application oriented})
\end{aligned}
\]
\end{document}